%% file: acl_latex.tex
\pdfoutput=1

\documentclass[11pt]{article}
\usepackage{amsmath} 

\usepackage[preprint]{acl}

\usepackage{times}
\usepackage{latexsym}
\usepackage{booktabs}
\usepackage{graphicx}
\usepackage{amsmath}
\usepackage{algorithm}
\usepackage{algpseudocode}
\usepackage{tcolorbox}
\usepackage{graphicx}
\usepackage{caption}
\usepackage{subcaption}
\usepackage{multirow}
\usepackage{booktabs}
\usepackage{multirow}
\usepackage{color}
\usepackage{array}
\usepackage{geometry}
\usepackage{mathtools}
\usepackage{multicol}

\DeclareUnicodeCharacter{2212}{-}

\newcommand{\rbnote}[1]{ {\textcolor{blue} { ***Rahul edit: #1}}}

\newcommand{\revisionnote}[1]{ {\textcolor{blue} {#1}}}
\usepackage[T1]{fontenc}

\usepackage[utf8]{inputenc}

\usepackage{microtype}

\usepackage{inconsolata}

\usepackage{graphicx}

\title{\textbf{\ours}: Ordering Examples On-The-Fly for In-Context Learning}

 \author{Rahul Atul Bhope \\ UC Irvine, USA 
         \And Praveen Venkateswaran \\ IBM Research AI, USA \And K. R. Jayaram \\ IBM Research AI, USA \AND Vatche Isahagian \\ IBM Research AI, USA \And Vinod Muthusamy \\ IBM Research AI, USA \And Nalini Venkatasubramanian \\ UC Irvine, USA
         }

\newcommand{\ours}{OptiSeq}
\newcommand{\ourstwo}{EOptiSeq}
\begin{document}
\maketitle

\begin{abstract}

Developers using LLMs and LLM-based agents in their applications have provided plenty of anecdotal evidence
that in-context-learning (ICL) is fragile.
In this paper, we show that in addition to the quantity and quality of examples, the order in which the in-context
examples are listed in the prompt affects the output of the LLM and, consequently, their performance. While prior work has explored improving ICL through dataset-dependent techniques, we introduce \ours, a purely inference-time, dataset-free optimization method that efficiently determines the best example order.
\ours\ leverages log probabilities of LLM-generated outputs to systematically prune the search space of possible orderings and recommend the best order(s) by distinguishing orderings that yield high levels of accuracy and those that underperform. Extensive empirical evaluation on
multiple LLMs, datasets, and prompts demonstrates that \ours\ improves accuracy by 5.5 - 10.5 \emph{percentage points} across multiple tasks.

\end{abstract}

\section{Introduction}\label{sec:intro}
\input{pv_intro}

\section{Sensitivity of ICL to Example Ordering}\label{sec:icexorder}

In this section, we present an analysis of the impact of in-context example ordering on performance under inference-time settings.

\begin{figure}[h]
    \centering
    \includegraphics[width=1\linewidth]{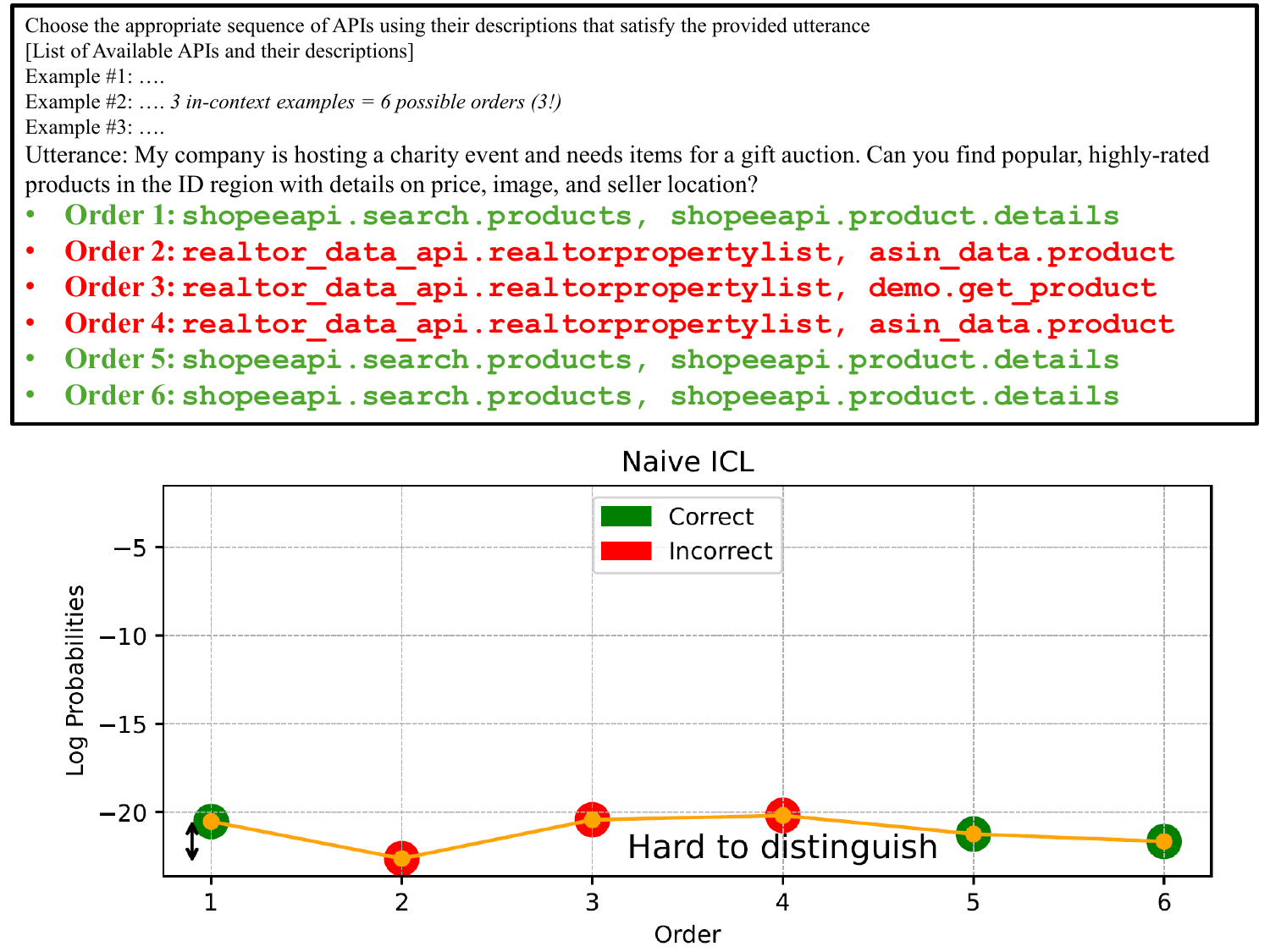}
    \caption{Naive ICL exhibits higher overlap in log probabilities for correct and incorrect outpus, making distinction harder.}

    \label{fig:model:log:variance:compare:1utt}
\end{figure}

\paragraph{\noindent Naive ICL fails to distinguish between correct and incorrect outputs.}
Figure~\ref{fig:model:log:variance:compare:1utt} illustrates an example from the ToolBench dataset, where Naive ICL is applied using the standard prompt structure. We evaluate all six in-context example permutations using \texttt{llama-3-8b-instruct} and compare their generated API sequences against the ground truth. Orders 1, 5, and 6 produce correct sequences, while Orders 2, 3, and 4 fail.
To analyze these failures, we compute the logarithmic probabilities of each output sequence as given by the LLM. In Figure~\ref{fig:model:log:variance:compare:1utt} we can see Naive ICL assigns similar log probabilities to both correct and incorrect sequences across all orders. Ideally, an LLM should assign higher probabilities to correct sequences, effectively distinguishing them from incorrect ones.

\begin{figure}[h]
    \centering
    \includegraphics[width=1\linewidth]{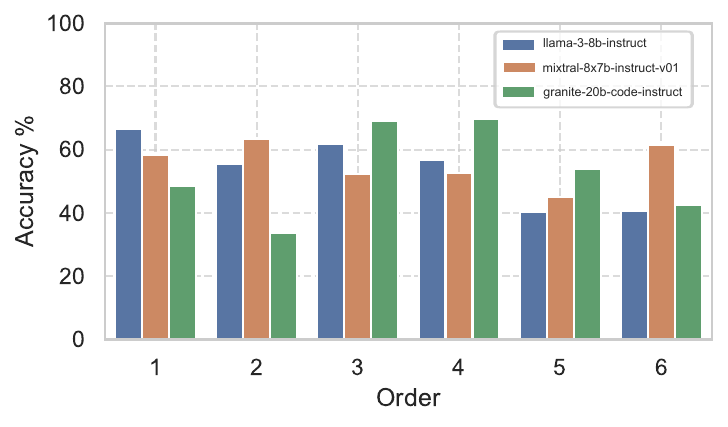}
    \caption{AG News classification accuracy across different orders and models }
    \label{fig:agnews:orders:models}
\end{figure}

\paragraph{\noindent\textbf{Optimal order varies across instances and models.}}
We evaluate six example orderings using three AG News~\cite{zhang2015character} articles as in-context examples across three models. Figure~\ref{fig:agnews:orders:models} shows that optimal ordering is not universal—it varies across test instances and models. A fixed order fails to generalize, necessitating an adaptive approach without relying on training data or precomputed orderings. Moreover, an order that performs well for one model (e.g., \texttt{llama-3-8b-instruct}) may underperform for another (e.g., \texttt{granite-20b-code-instruct}), highlighting the need for a model-aware, instance-specific ordering strategy without assuming dataset-wide fairness or requiring labeled data.

\paragraph{\noindent
\textbf{Impact of choosing the wrong order can be significant depending on the dataset or task.}}

\begin{figure}[h]
\centering
\includegraphics[width=1\linewidth]{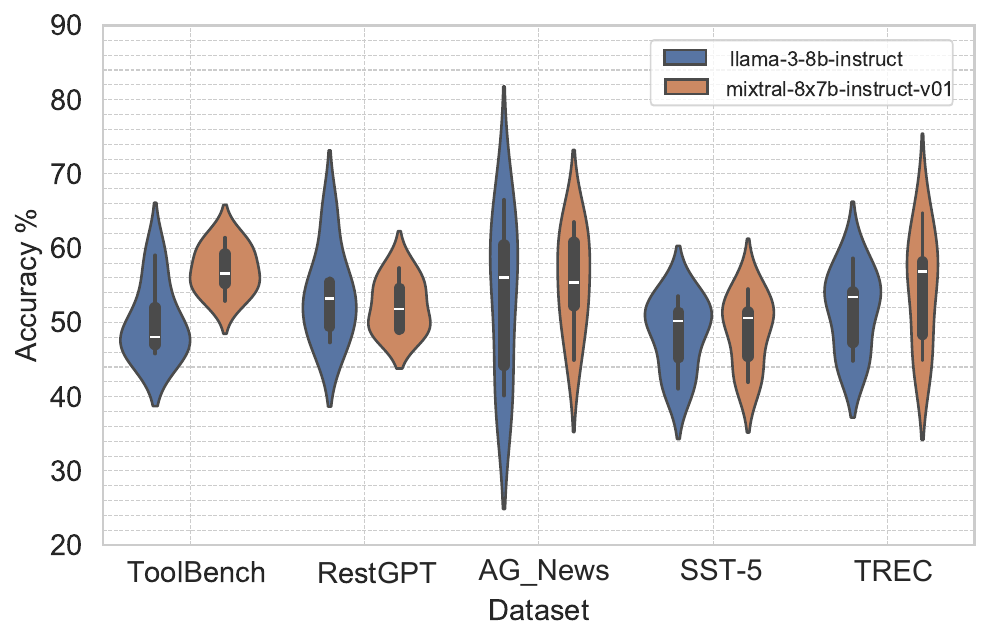}
\caption{Variation in the average accuracy across all permutations of three examples for five datasets}
\label{fig:motivation:orders}
\end{figure}

We evaluate five datasets (Section~\ref{sec:datasets}) across two tasks—API sequence generation and text classification—by measuring accuracy variations across all possible example orderings in a three-shot setting. Figure~\ref{fig:motivation:orders} demonstrates that example ordering plays a crucial role in performance, with accuracy fluctuating by $\approx 12$ percentage points for API generation and $\approx 17$ percentage points for classification. This suggests that the model is influenced by example order than by the examples themselves.

\input{methodology}

\section{Evaluation}\label{sec:evaluation}

We evaluate our proposed methodologies \ours\ and \ourstwo\ across two tasks, five datasets, and five Large Language Models across three LLM families. To ensure consistency across all experiments, we utilize greedy decoding and implement 3-shot ICL for all models and datasets. This 3-shot approach allows for efficient batching of LLM inferences for permutations with in-context examples, while balancing the trade-off between larger context sizes and increased latency, which is crucial for inference-time applications. Details of our experimental setup are provided below.

\subsection{Datasets}\label{sec:datasets}

We evaluate our approach on \emph{two} different tasks: API sequence generation and text classification. API generation requires the model to generate a sequence from a set of API candidates (i.e.) multi-label prediction, resulting in a large combinatorial solution space that prior approaches do not address~\cite{guo2024makes,lu2021fantastically}. For text classification we use (i) AG News~\citep{zhang2015character}, (ii) SST5~\citep{socher2013recursive}, and (iii) TREC~\citep{hovy2001question}, while for the API generation task we use (iv) RestGPT~\citep{wu2023restgpt}, and (v) ToolBench~\citep{lu2022toolbench}.

\subsection{Models} 
We evaluate these datasets across five models from three model families with a diverse range of parameters ranging from $8$B to $70$B --
(i) \texttt{llama-3-8b-instruct} and (ii) \texttt{llama-3-70b-instruct}~\citep{touvron2023llama3}, (iii) \texttt{granite-13b-instruct-v2} and (iv) \texttt{granite-20b-code-instruct}~\citep{ibm2023granite}, and (v) \texttt{mixtral-8x7b-instruct-v01}~\citep{mistral2023mixtral}, which uses the mixture of experts approach. This heterogeneity in model parameters and architecture allows for a comprehensive assessment of performance across varying scales of computational complexity and capability.

\subsection{Comparative Techniques} 
We compare \ours\ and \ourstwo\ against random order selection and Top-$k$ order selection.  In random selection, an order is selected at random for each test instance. In Top-$k$, the cosine similarity is calculated between each in-context example and the test instance, and the examples are arranged in decreasing order based on their similarity scores. 


We also compare against recent baselines LocalE~\cite{lu2021fantastically} and Influence score~\cite{guo2024makes}. LocalE computes output token probabilities from the first ICL task in \ours\ and their entropy for each example order \( \phi \): \( Ent(\phi) = - \sum_{y} P(y | C_{\phi}) \log P(y | C_{\phi}) \), selecting the order with median entropy to balance model confidence. The Influence score measures each order's deviation from expected probability: \( I(x_t, \phi) = P(y | x_t, C_{\phi}) - \frac{1}{|\Phi|} \sum_{\phi' \in \Phi} P(y | x_t, C_{\phi'}) \), capturing the ordering's relative impact on prediction. However, these methods rely on corpus-level properties: LocalE needs label fairness assumptions and artificial development sets, while Influence score assumes implicit and fair label distribution across orderings. In contrast, \ours\ and \ourstwo\ operate without validation datasets, performing purely inference-time optimization without corpus-level assumptions, making direct comparisons challenging.

Additionally, we focus on example ordering, not selection, ensuring all techniques operate on identical in-context examples to isolate ordering effects. This approach avoids unfair comparisons that would arise from different example sets. \ours\ optimizes ordering using log probabilities, independent of specific examples or dataset heuristics, ensuring broad applicability and consistency across various example sets.


\subsection{Metrics} For the classification tasks, we report the Accuracy in \% for the entire dataset.
For the API sequence generation task, we report the following metrics:

\begin{itemize}
    \item Accuracy: Represents the fraction of test cases where the generated API sequence exactly matches the ground truth (in correct order), compared to the ground truth sequence.
    \item Recall: For each test utterance, this metric represents the fraction of correctly predicted APIs (ignoring order) compared to the total number of APIs in the ground truth.
    \item Precision: For each test utterance, this metric represents the fraction of correctly predicted APIs (ignoring order) compared to the total number of APIs in the predicted sequence.
\end{itemize}

\section{Results and Discussions}~\label{sec:eval}

\input{results}

\section{Related Work}

\textbf{Methods for Optimizing Example Ordering:}~\cite{xu2024label_distributions} formulates example ordering as an optimization problem. Using label proportion, it improves accuracy and reduces model miscalibration across classification tasks.~\cite{zhang2024batch_icl} Batch-ICL aggregates meta-gradients from independent computations, making the model agnostic to example order while improving performance and reducing computational costs.~\cite{wu2022self_adaptive} Proposes a select-then-rank framework for self-adaptive ICL, achieving significant performance gains by dynamically optimizing example orders.
Inspired by how humans learn,~\cite{liu2024curriculum} gradually increases example complexity, improving instance and corpus-level performance through curriculum ordering. Unlike batch or curriculum-based approaches, \ours\ performs instance-specific optimization rather than applying a general rule.

\textbf{Example Selection and Ranking Techniques: } \cite{gupta2023coverage} Selects diverse and informative examples using BERTScore-Recall, significantly outperforming independent ranking methods. DEmO \cite{guo2024makes} identifies optimal example orders for individual instances through label fairness and content-free metrics. \cite{liu2024se2} formulates example selection as a sequential process using beam search to optimize inter-relationships and diversity among examples. EXPLORA \cite{purohit2024explora} improves task-specific exemplar selection for complex reasoning tasks by efficiently estimating scoring function parameters, reducing computational cost while enhancing performance. CEIL \cite{ye2023compositional} models example selection as a subset selection problem using Determinantal Point Processes and contrastive learning to optimize example interactions across diverse NLP tasks. \ours\ goes beyond static or sequential ranking by dynamically testing every possible example order and using log probabilities to determine the best sequence.

\textbf{Theoretical Insights and Adaptive Strategies in ICL}
~\cite{chandra2024predicting} demonstrates that dynamically adjusting the number of in-context examples improves task-specific performance over fixed hyperparameters.
~\cite{zhao2024instruction_following} examines the limitations of ICL for instruction-following tasks and identifies key parameters for alignment.
~\cite{do2023adversarial_icl} employs adversarial learning to iteratively refine prompts, significantly improving performance across diverse tasks. \ours\ avoids the complexity of adversarial learning or fine-tuning, providing an inference-time method for ICL.

\section{Conclusion}

This study highlights the impact of in-context example ordering on LLM performance. \ours\ significantly enhances model accuracy by optimizing example orderings. By evaluating all possible orderings and selecting the highest confidence score based on input log probabilities, \ours\ consistently improved accuracy by 5.5 to 10.5 percentage points over baselines. This improvement was observed across various API sequence generation and text classification tasks for three different model families with diverse parameter ranges, demonstrating \ours's robustness and versatility in enhancing ICL.

\section{Limitations}


\ours\ achieves better results than Top-K and Random order selection but requires evaluating $k = |\mathcal{E}|!$ permutations of prompts, which introduces computational challenges as the number of in-context examples grows. Our experiments confirm that fixed example orderings struggle to generalize across tasks, instances, and model architectures (Section~\ref{sec:icexorder}). This limitation arises from the strong dependence between the optimal ordering, the characteristics of the examples, and the model-specific biases. Additionally, our work focuses on instance-specific adaptive ordering, which optimizes example sequences for individual inputs. While this approach maximizes performance for a given instance, we recognize that it does not inherently address cross-instance or cross-model generalization. A promising future direction is exploring methods using meta-learning, or domain adaptation to learn transferable ordering strategies that can be applied across various instances and models without repeated optimization. Furthermore, while the approach relies on logarithmic probability evaluations for optimal permutation selection, not all LLM platforms and APIs services currently support token-level log-probability computation. However, as models continue to evolve and LLM platforms expand to include more granular scoring features, the applicability and efficiency of \ours\ are likely to improve, paving the way for broader adoption in real-world scenarios.



\bibliography{custom}
\clearpage
\appendix

\input{appendix}

\end{document}

%% file: pv_intro.tex
The use of in-context learning (ICL) with large language models (LLMs) has become a popular approach to achieve impressive performance in many NLP tasks \citep{raffel2020exploring, radford2019language}. 
In ICL, models are prompted during inference with task-specific examples that help condition the generated output. Unlike fine-tuning, it does not require updates to the model parameters, which offers many benefits with ever-increasing model sizes and capabilities \citep{brown2020language}.
It has been shown that prompting LLMs without fine-tuning is often sensitive to prompt design \citep{shin2020autoprompt, chen2022meta}. In particular, the quality, quantity, and permutation of examples can all significantly impact performance \citep{zhao2021calibrate}. To address this, previous work has primarily focused on selecting high-quality examples from a candidate set \citep{yang-etal-2023-representative, liu2021makes} and determining the optimal number of these examples \citep{zhang2023makes, agarwal2024many}.

\begin{figure*}[h]
    \centering
    \includegraphics[width=1\linewidth]{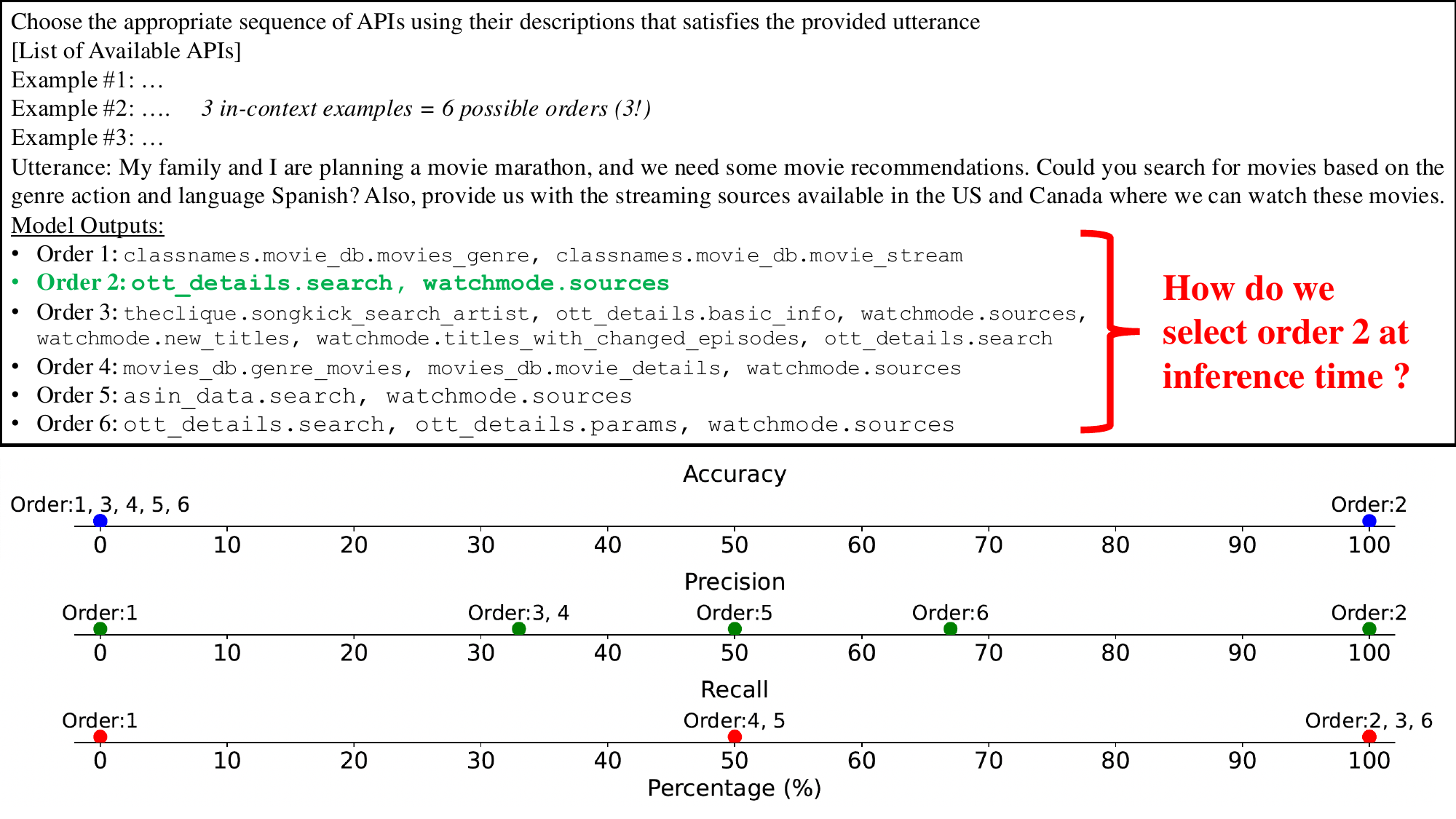}
    \caption{\texttt{llama-3-8b-instruct} performance variation with 6 in-context example orderings using 3 examples.}
    \label{fig:api:seq:gen:motivation}
\end{figure*}

Existing solutions to mitigate prompt sensitivity caused by example ordering at \emph{inference-time} are limited. Figure \ref{fig:api:seq:gen:motivation} illustrates this using an API sequence generation task on the ToolBench dataset \citep{lu2022toolbench}. Given three in-context examples, LLM predictions vary significantly across the six possible orderings, with only one specific order (\textit{order 2}) yielding the correct answer. This variability in precision and recall underscores how reordering examples alters the input context, influencing token probabilities and ultimately affecting model performance. Most prior approaches rely on a precomputed strategy that assumes access to a predefined set of examples and a fixed label space~\cite{lu2021fantastically, guo2024makes}. However, our setting is inherently online, requiring dynamic ordering decisions at inference time without training or validation data. Since orderings cannot be precomputed, the limited number of examples further complicates the problem.

A common approach to selecting examples at inference-time is to generate embeddings of candidate examples using a model like Sentence-BERT \cite{reimers2019sentence} and retrieve the top-$k$ most similar examples for a given test instance, ranking them based on distance or similarity.
However, \textbf{there is a distinction between ranking examples (determining how relevant they are to our test case) and ordering them (deciding how to arrange them in the prompt)}. While finding relevant examples through ranking is valuable, it does not tell us the best way to order them in the prompt. Furthermore, top-$k$ is dependent on
the quality of embeddings and can lead to suboptimal performance if the distances are too close.
Recent efforts leverage additional in-domain validation datasets to perform offline evaluation of different orders, which is often not feasible in real-world scenarios where limited data is available \citep{perez2021truefewshotlearninglanguag}. 
They also develop mutual information- or entropy-based heuristics \citep{sorensen2022information, lu2021fantastically,guo2024makes} on these validation datasets, but are computationally limited to tasks like single-class classification assuming label-balanced tasks and do not generalize to generation tasks.
Moreover, as we show in Section \ref{sec:icexorder}, the optimal order of the examples varies for the test samples within and between tasks and between different LLMs, 
finding and selecting this optimal order is very challenging in production settings. 

An effective real-world solution needs to be (a) non-reliant on the availability of additional validation/training data, (b) generalizable to different tasks, and (c) performant across different LLMs and number of examples.
In this paper, we introduce \texttt{\ours}, a novel approach for selecting the optimal order of in-context examples at run-time and make the following contributions:



\begin{itemize}
    \item We present a study of \emph{example-order} sensitivity in ICL in an inference-time setting (Section~\ref{sec:icexorder}).
    
    \item We describe the design and implementation of \ours, which evaluates few-shot ICL performance across order permutations and then selects the best order by leveraging the model's ability to distinguish between outputs (Sections~\ref{sec:core},~\ref{sec:eval}).

    \item We propose \ourstwo, a variant of \ours\ that evaluates fewer permutations at inference time, achieving lower accuracy gains but improving efficiency (Sections~\ref{sec:core},~\ref{sec:eval}).
    

    \item We present a detailed empirical evaluation of \ours\ and \ourstwo\  (Section~\ref{sec:eval}), 
    across two tasks: API sequence generation and classification, five datasets, and five LLMs (8B–70B parameters). \ours\ improves accuracy by 10.5 percentage points over random selection, 6.5 percentage points over Top-K selection and 5.5 percentage points over recent baselines~\cite{lu2021fantastically, guo2024makes}.

\end{itemize}


%% file: methodology.tex
\section{Methodology }~\label{sec:core}

The effectiveness of in-context learning in large language models (LLMs) depends on how well the model utilizes contextual cues, which in turn is influenced by the ordering of examples.  Section~\ref{sec:icexorder} demonstrates that the optimal order is shaped by both the characteristics of the examples and the model itself, making it difficult to predict solely from the inputs. This leads to a key question:
\textit{How does the ordering of in-context examples affect an LLM’s ability to distinguish between possible outputs?} The ordering of examples provides a signal that influences the LLM’s predictions, often leading the model to generate different outputs with comparable log probabilities across different orders as seen in Figure~\ref{fig:model:log:variance:compare:1utt}. This suggests that LLMs treat the entire prompt—content and order—as a holistic sequence, generating the output they are most confident in given that specific sequence. If we can evaluate the generated output in a way that removes the influence of ordering context, we would be able to better distinguish which outputs are inherently more likely to be correct, independent of example order.

\subsection{\ours}

\begin{figure*}[h]
    \centering
    \includegraphics[width=1\linewidth]{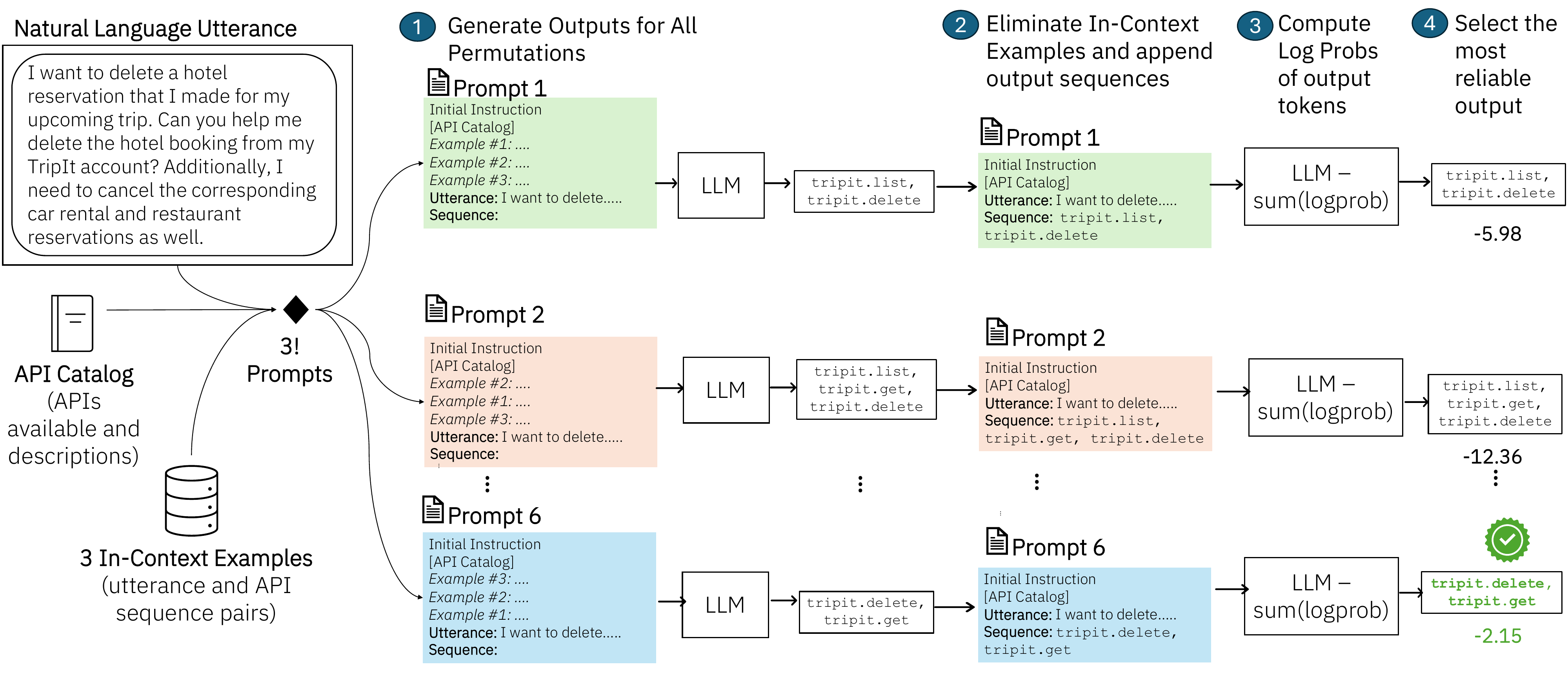}
    \caption{OptiSeq Overview}
    \label{fig:optiseq:overview}
\end{figure*}

\begin{figure}[h]
    \centering
    \subfloat[\vspace{-2mm}]{{\includegraphics[width=0.22\textwidth]{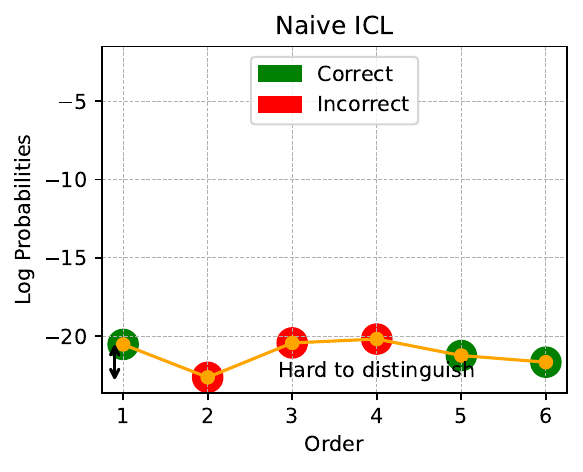}}%
\label{fig:log_probs_wo_OptiSeq}
    }
    \quad
    \subfloat[\vspace{-2mm}]{\includegraphics[width=0.22\textwidth]{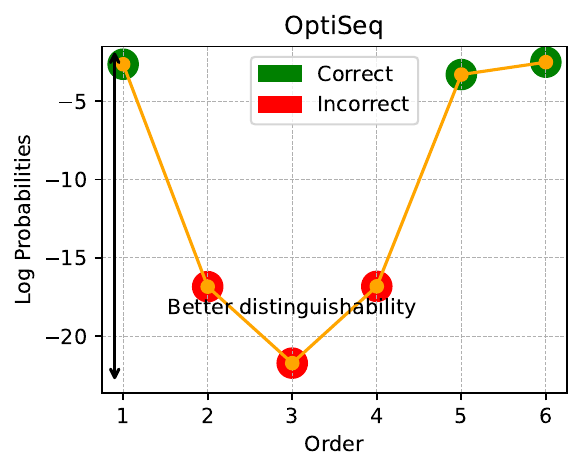}%
\label{fig:log_probs_with_OptiSeq}
    }  
    \caption{Comparison of log probabilities with and without OptiSeq optimization}
    \label{fig:log_probs_comparison}
\end{figure}

Building on this insight, we introduce \texttt{\ours}, an \textit{example-free} approach that optimizes in-context example ordering by leveraging the log probabilities of LLM-generated outputs (Figure~\ref{fig:optiseq:overview}). The process consists of the following steps:
\begin{itemize}
    \item \textbf{Generate  outputs for all permutations:}  
    Given a task instruction $\mathcal{I}$ with in-context examples $\mathcal{E}$, we construct $k = |\mathcal{E}|!$ prompts, each corresponding to a unique ordering of examples. These prompts are fed into the LLM to generate candidate outputs $o_k \in \mathcal{O}$ given by:
\begin{equation} \label{eq:naive:icl}
    \small
   \log P(o_k \;|\; \mathcal{I}, \mathcal{E}_k) =
   \sum_{i=1}^{n} \log P(x_{ik} \;|\; \mathcal{I} \oplus \mathcal{E}_k \oplus x_{jk<ik})
\end{equation}
where $x_{ik}$ is the $i^{th}$ token of output $o_k$,  $\mathcal{E}_k$ is the $k^{th}$ example permutation and $x_{jk<ik}$ represents preceding tokens providing autoregressive context.
    \item \textbf{Eliminate In-context examples:}  
    For each candidate output, we modify the prompt by removing the in-context examples while retaining only the task instructions $\mathcal{I}$.
    \item \textbf{Append candidate outputs:}  
    Each generated output $o_k$ is appended to its corresponding modified prompt.
    \item \textbf{Compute log probabilities:}  
    Using the same LLM, we compute the sum of the log probabilities of the output tokens $o_k$, conditioned only on the task instructions:
    \begin{align*}  
        \Phi_k &= \sum_{i=1}^{n} \log P(x_{ik} \;|\; \mathcal{I} \oplus x_{jk<ik})  \forall o_k \in \mathcal{O} \\
    \end{align*}
    \item \textbf{Select the optimal ordering:}  
    The order $k^*$ that maximizes the sum of log probabilities is chosen as the optimal in-context example ordering.
    \begin{align*}
        k^* &= \underset{k}{\mathrm{argmax}}\, \Phi_k 
    \end{align*}
\end{itemize}
This results in better distinguishability among outputs at inference-time as seen in Figure~\ref{fig:log_probs_comparison} for the same utterance from Figure~\ref{fig:model:log:variance:compare:1utt}. The full algorithm is demonstrated in Algorithm~\ref{alg:ours}. 

\begin{algorithm}[t]
\caption{\texttt{\ours}}
\label{alg:ours}
\small
\textbf{Inputs:} Task instruction $\mathcal{I}$, In-context examples $\mathcal{E}$, Large Language Model $M$\\ 
\textbf{Outputs:} Optimal example ordering $\mathcal{E}_{k^*}$
\begin{algorithmic}[1]
\State Construct $k = |\mathcal{E}|!$ permutations $\mathcal{E}_k$ 
\For{each permutation $\mathcal{E}_k$}
    \State $o_k \gets M(\mathcal{I}, \mathcal{E}_k)$ \textcolor{blue}{\footnotesize// Generate outputs for all orderings}
\EndFor

\For{each generated output $o_k$}
    \State $\Phi_k \gets \sum_{i=1}^{n} \log P(x_{ik} \mid \mathcal{I} \oplus x_{jk<ik})$ \textcolor{blue}{\footnotesize// Compute log probs without examples}
\EndFor

\State $\begin{aligned} k^* &= \underset{k}{\mathrm{argmax}}\, \Phi_k \end{aligned}$ \textcolor{blue}{\footnotesize// Select optimal ordering}

\State \Return $\mathcal{E}_{k^*}$
\end{algorithmic}
\end{algorithm}

\subsection{\ourstwo\ }

While \ours\ enhances distinguishability by computing log probabilities without in-context examples, it still requires evaluating multiple example orderings to find the most effective one. However, exhaustively searching over all $|\mathcal{E}|!$ permutations is computationally costly for a higher number of examples. To prune the permutation search space during inference-time we propose \ourstwo\ .


\ourstwo\ optimizes in-context example ordering at inference time without exhaustive search of all $|\mathcal{E}|!$ permutations. Given $x_{\text{test}}$ and examples $\{e_1, \dots, e_E\}$, it computes embeddings via Sentence-BERT~\cite{reimers2019sentence}:
\begin{equation}
   \mathbf{v}_i = \text{SBERT}(e_i), \quad \mathbf{v}_{\text{test}} = \text{SBERT}(x_{\text{test}})
\end{equation}
and calculates cosine similarities:
\begin{equation}
   s_i = \frac{\mathbf{v}_i \cdot \mathbf{v}_{\text{test}}}{\|\mathbf{v}_i\| \|\mathbf{v}_{\text{test}}\|}
\end{equation}
The top-$\mathcal{E}$ examples are selected and the highest-ranked one based on cosine-similarity is anchored first, requiring only $(\mathcal{E}-1)!$ permutations for the remaining examples. This approach reduces evaluations from $\mathcal{E}!$ to $(\mathcal{E}-1)!$. This strategy is inspired by ~\cite{liu2024lost}, which shows that placing the most similar (highest-contextual-relevance) example in the first position results in the highest accuracy. The final ordering is selected using \textit{example-free} log prob computation as in \ours.

%% file: results.tex
\begin{table*}[htp]
\centering
\small	
\renewcommand{\arraystretch}{1.05} 
\setlength{\tabcolsep}{6pt} 
\begin{tabular}{c|l|cccccc}
\hline
\textbf{Dataset} & \textbf{Model} & \textbf{Random} & \textbf{Top-K} & \textbf{OptiSeq} & \textbf{EOptiSeq} & \textbf{LocalE} & \textbf{Influence} \\
\hline
\multirow{5}{*}{Tool Bench} 
    & llama-3-8b-instruct      & 43.99 & 44.80 & \textbf{54.18} & \underline{51.12} & 48.77 & 50.30 \\
    & llama-3-70b-instruct     & 58.24 & 59.06 & \textbf{68.43} & \underline{65.37} & 63.03 & 64.56  \\
    & granite-20b-code-instruct       & 53.76 & 56.21 & \textbf{62.93} & 60.28 & 60.32 & \underline{61.54} \\
    & granite-13b-instruct-v2         & 43.42 & 44.41 & \textbf{47.76} & \underline{46.37} & 43.31 & 44.64 \\
    & mixtral-8x7b-instruct-v01 & 43.38 & 42.56 & \textbf{48.26} & 46.23 & \underline{46.90} & 42.26 \\ \hline
\multirow{5}{*}{RestGPT}    
    & llama-3-8b-instruct      & 40.76 & 42.04 & \textbf{61.15} & \underline{52.32} & 48.46 & 51.09  \\
    & llama-3-70b-instruct     & 54.14 & 57.96 & \textbf{69.42} & 64.34 & 61.48 & \underline{64.97}  \\
    & granite-20b-code-instruct       & 35.03 & 39.49 & \textbf{46.49} & \underline{42.43} & 38.15 & 38.15 \\
    & granite-13b-instruct-v2         & 24.84 & 24.20 & \textbf{30.57} & \underline{28.02} & 25.87 & 26.71 \\
    & mixtral-8x7b-instruct-v01 & 41.41 & 43.31 & \textbf{55.41} & \underline{51.59} & 43.17 & 42.15 \\ \hline
\multirow{5}{*}{AGNews}     
    & llama-3-8b-instruct      & 72.13 & 73.89 & \textbf{78.54} & 75.64 & 74.44 & \underline{76.53} \\
    & llama-3-70b-instruct     & 75.00 & 76.50 & \textbf{81.00} & 78.00 & 77.50 & \underline{79.03} \\
    & granite-20b-code-instruct       & 62.90 & 66.31 & \textbf{73.94} & \underline{69.27} & 61.99 & 65.92 \\
    & granite-13b-instruct-v2         & 59.81 & 58.69 & \textbf{61.36} & 59.98 & 59.42 & \underline{60.52} \\
    & mixtral-8x7b-instruct-v01 & 85.45 & 85.97 & \textbf{89.60} & 86.37 & 86.78 & \underline{87.62}\\ \hline
\multirow{5}{*}{SST-5}      
    & llama-3-8b-instruct      & 53.10 & 53.10 & \textbf{57.10} & 54.10 & \underline{54.87} & 54.00 \\
    & llama-3-70b-instruct     & 55.00 & 55.50 & \textbf{60.00} & 56.00 & \underline{57.20} & 56.80 \\
    & granite-20b-code-instruct       & 27.80 & 30.50 & \textbf{32.70} & \underline{30.70} & 28.66 & 29.16 \\
    & granite-13b-instruct-v2         & 46.90 & 47.90 & \textbf{50.10} & \underline{48.90} & 47.15 & 47.45 \\
    & mixtral-8x7b-instruct-v01 & 52.90 & 53.90 & \textbf{55.70} & \underline{54.90} & 54.38 & 54.85 \\ \hline
\multirow{5}{*}{TREC}       
    & llama-3-8b-instruct      & 60.60 & 63.20 & \textbf{67.80} & \underline{66.40} & 63.80 & 64.20 \\
    & llama-3-70b-instruct     & 62.00 & 65.00 & \textbf{70.00} & \underline{68.00} & 65.60 & 66.00 \\
    & granite-20b-code-instruct       & 50.60 & 53.40 & \textbf{58.00} & \underline{55.60} & 50.98 & 51.34 \\
    & granite-13b-instruct-v2         & 40.00 & 43.20 & \textbf{46.40} & \underline{45.20} & 40.36 & 40.69 \\
    & mixtral-8x7b-instruct-v01 & 62.20 & 66.80 & \textbf{73.20} & \underline{70.60} & 68.94 & 69.40 \\ \hline
\end{tabular}
\caption{Updated accuracy (\%) comparison of various models across datasets. Bold values indicate the highest accuracy, and underlined values represent the second-highest accuracy.}
\label{table:main:results}
\end{table*}

\textbf{\ours\ shows improved performance over Top-K and random selection}. Table~\ref{table:main:results} highlights experimental results for different tasks. \ours\ achieves an average improvement of 10.5\% points over random selection, 9.05\% points over Top-K, 6.5 \% over LocalE and 5.5 \% over Influence score for the API sequence generation task. For text classification task, \ours\ demonstrates an approximate improvement of 6\% points compared to random selection and Top-K and 4 \% over LocalE and Influence score. LocalE and Influence take into account example ordering while measuring log probability, which reduces distinguishability among order permutations. \ours\ evaluates all permutations and then analyzes the output sequences in an \textit{example-free} setting, which leads to performance improvements due to better distinguishability. \ourstwo, which builds on principles of \ours\ and Top-K, performs marginally better than Top-K but worse than \ours. This is attributed to the fact that it evaluates fewer permutations than \ours\ but still uses zero-shot inference to improve ICL.

\begin{figure} [h]
    \centering
    \includegraphics[width=1\linewidth]{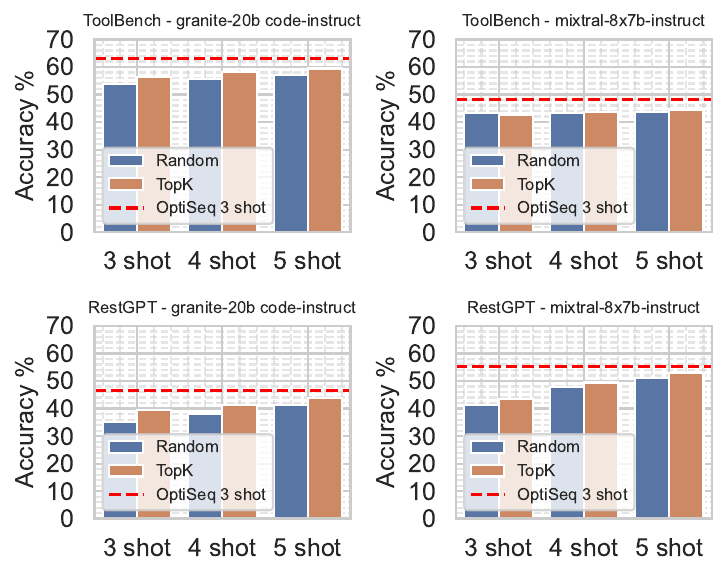}
    \caption{Comparing 3, 4, and 5 shot random/Top-K selection strategy with 3 shot \ours\ (in red) .}
    \label{fig:345:shot:optiseq}
\end{figure}

\textbf{The strategic ordering of a smaller number of examples in \ours\ can significantly enhance performance compared to using a larger set of examples in a random/Top-K order}. As highlighted in Figure~\ref{fig:345:shot:optiseq}, ordered 3 shot ICL using \ours\ performs better than 4 and 5 shot ICL using a random order and Top-K. On average, for the API sequence generation task, \ours\ performs better than random and top-k selection by 5.07\% points and 2.1\% points for classification task (shown in Figure~\ref{fig:345:shot:optiseq:class}). Adding more examples does not guarantee better performance, especially given context-length limitations. Exceeding the model's input window can lead to prompt truncation and degraded performance at inference-time. ~\cite{liu2023pre} demonstrates that more examples may introduce noise or redundancy, limiting generalization. \ours\ offers a robust solution for ICL by focusing on order optimization, which remains effective even when adding more examples is infeasible.

\begin{figure} [h]
    \centering
    \includegraphics[width=1\linewidth]{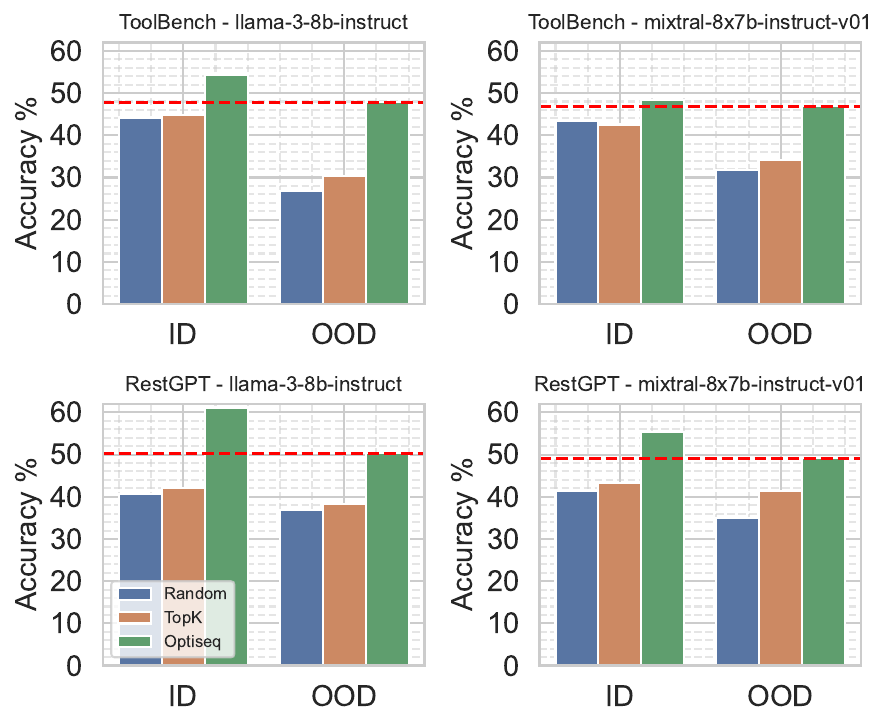}
    \caption{Comparing In-distribution with Out-of-distribution performance. Here ToolBench uses in-context examples from RestGPT and vice-versa. Results shown for 2 models.}
    \label{fig:ood:optiseq}
\end{figure}

\textbf{The strategic ordering of out-of-distribution (OOD) examples using \ours\ can lead to better performance compared to using in-distribution (ID) examples in a random/Top-K order}. We evaluate the 3 shot ICL API sequence generation task for ID and OOD examples. Here, the ToolBench dataset uses examples from RestGPT, and RestGPT uses examples from ToolBench. Figure~\ref{fig:ood:optiseq} shows that the performance drops when we use OOD examples across all techniques. The drop is significantly high for random and Top-K selection. On average \ours\ using OOD examples performs better than Top-K using ID examples by 5.35\% points and random selection using ID examples by 6.15\% points.

\begin{figure}[h]
    \centering
    \begin{subfigure}{0.23\textwidth}
        \centering
        \includegraphics[width=\textwidth]{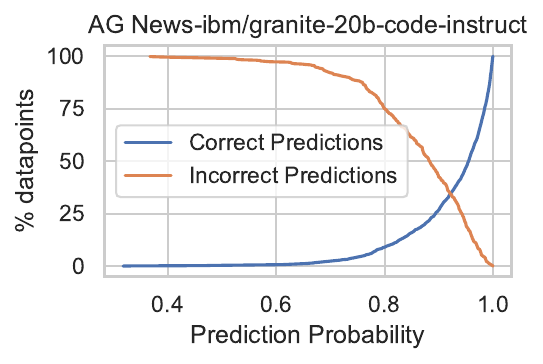}
        \label{fig:plot1}
    \end{subfigure}
    \hfill
    \begin{subfigure}{0.23\textwidth}
        \centering
        \includegraphics[width=\textwidth]{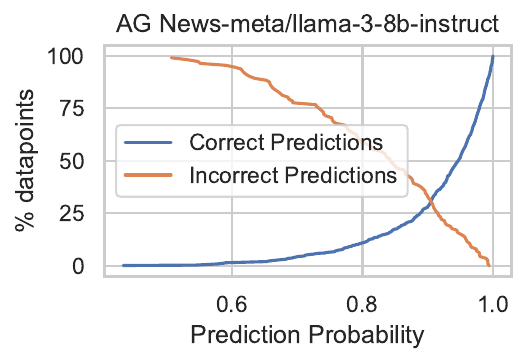}
        \label{fig:plot2}
    \end{subfigure}
    \begin{subfigure}{0.23\textwidth}
        \centering
        \includegraphics[width=\textwidth]{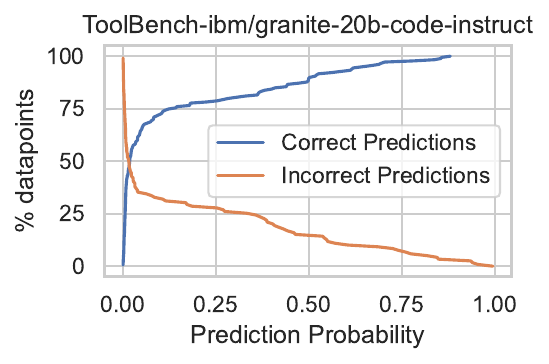}
        \label{fig:plot3}
    \end{subfigure}
    \hfill
    \begin{subfigure}{0.23\textwidth}
        \centering
        \includegraphics[width=\textwidth]{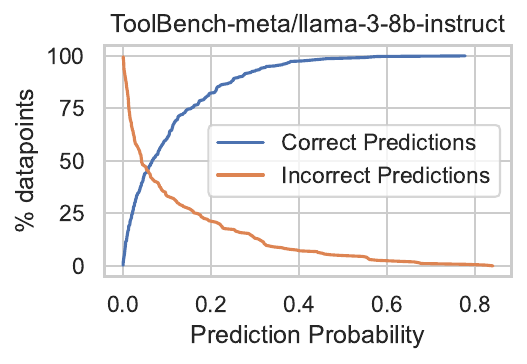}
        \label{fig:plot4}
    \end{subfigure}
    \caption{Probability distribution for correct and incorrect predictions, reflecting the model's confidence.}
    \label{fig:four_plots:distri}
\end{figure}



\begin{table*}[!ht]
\centering
\small
\renewcommand{\arraystretch}{1.05} 
\setlength{\tabcolsep}{2.5pt} 
\begin{tabular}{c|l|cc|cc|cc|cc|cc|cc}
\hline
\textbf{Dataset} & \multicolumn{1}{c|}{\textbf{Model}} & \multicolumn{2}{c|}{\textbf{Random}} & \multicolumn{2}{c|}{\textbf{TopK}} & \multicolumn{2}{c|}{\textbf{OptiSeq}} & \multicolumn{2}{c|}{\textbf{EOptiSeq}} & \multicolumn{2}{c|}{\textbf{LocalE}} & \multicolumn{2}{c}{\textbf{Influence}} \\
 &  & Prec. & Rec. & Prec. & Rec. & Prec. & Rec. & Prec. & Rec. & Prec. & Rec. & Prec. & Rec. \\ \hline
Tool Bench & llama-3-8b-instruct & 71.47 & 70.58 & 71.63 & 70.45 & 77.75 & 77.12 & 75.77 & 74.58 & 74.33 & 73.14 & 75.61 & 74.18  \\
 & llama-3-70b-instruct & 82.57 & 82.08 & 83.04 & 82.95 & 85.03 & 84.07 & 84.79 & 83.88 & 83.68 & 82.72 & 84.23 & 83.11 \\
 & granite-20b-code-instruct & 75.83 & 75.18 & 77.35 & 77.10 & 81.78 & 81.64 & 80.72 & 79.34 & 81.07 & 80.16 & 81.31 & 79.96 \\
 & granite-13b-instruct-v2 & 72.03 & 74.85 & 73.20 & 76.29 & 76.55 & 78.58 & 75.16 & 77.04 & 69.72 & 74.40 & 71.18 & 74.85 \\
 & mixtral-8x7b-instruct-v01 & 74.16 & 76.15 & 73.78 & 76.33 & 81.85 & 84.21 & 80.12 & 81.97 & 79.28 & 80.99 & 67.72 & 68.24 \\ \hline
RestGPT & llama-3-8b-instruct & 75.43 & 75.04 & 75.74 & 75.62 & 78.69 & 78.72 & 77.52 & 77.11 & 75.59 & 75.33 & 76.68 & 75.97 \\
 & llama-3-70b-instruct & 74.70 & 75.35 & 74.22 & 75.15 & 77.64 & 78.07 & 76.79 & 77.13 & 75.18 & 75.23 & 76.43 & 77.45 \\
 & granite-20b-code-instruct & 74.76 & 74.23 & 75.28 & 74.84 & 76.27 & 76.01 & 75.86 & 74.79 & 75.51 & 75.01 & 73.71 & 72.77 \\
 & granite-13b-instruct-v2 & 74.12 & 74.56 & 74.68 & 75.12 & 76.23 & 76.78 & 75.89 & 76.34 & 73.56 & 73.71 & 74.39 & 71.97 \\
 & mixtral-8x7b-instruct-v01 & 74.56 & 74.84 & 75.24 & 75.76 & 77.53 & 77.47 & 77.10 & 77.32 & 75.51 & 76.01 & 73.71 & 72.77 \\ \hline
\end{tabular}
\caption{Precision (Prec.) and Recall (Rec.) in \% for different models and datasets.}
\label{table:prec:rec}
\end{table*}

\textbf{\ours\ enhances accuracy using confidence metrics}. Figure~\ref{fig:four_plots:distri} shows that correct predictions generally have higher probabilities, indicating greater model confidence. The distribution of correct predictions skews towards higher confidence levels, while incorrect predictions tend to have lower probabilities. This pattern demonstrates the potential of using confidence metrics to improve model accuracy, possibly by filtering or adjusting predictions based on their confidence levels.

\textbf{\ours\ improves instance level predictions for API sequence generation}. Table~\ref{table:prec:rec} shows the  Precision and Recall metrics of instance level predictions at run-time for API sequence generation task. \ours\ and \ourstwo\ achieve an average improvement of 3.52\% in Precision and 3.21\% points in Recall over Random or Top-K, respectively and 3.01 \% in Precision and 3.37 \% in Recall over recent baselines. This indicates that our approaches induce the inclusion of more relevant APIs in the sequence compared to baselines.



\textbf{\ours\ Overheads: } \ours\ uses batched inference to reduce evaluation latency. To assess the efficiency of batched inference, we compared single and batched inference times on an NVIDIA A5000 GPU averaged across 100 runs for each setting (single and batched) for AG News using \texttt{llama-3-8b-instruct}. Initial runs were discarded for warm-up to mitigate initialization overhead and early measurement variability. Single Prompt Inference: 13.44s per prompt. Batched Inference (6 or 3! Prompts): 16.87s for the full batch. Batching significantly reduces per-prompt latency (from 13.44s to 2.81s), making inference more efficient. This reduces the latency of OptiSeq to $\approx2$ sequential single-prompt LLM calls.

 

%% file: appendix.tex
\section{Appendix: Additional Figures}
\label{sec:appendix}

    \begin{figure}[htbp]
    \centering
    \includegraphics[width=0.90\linewidth]{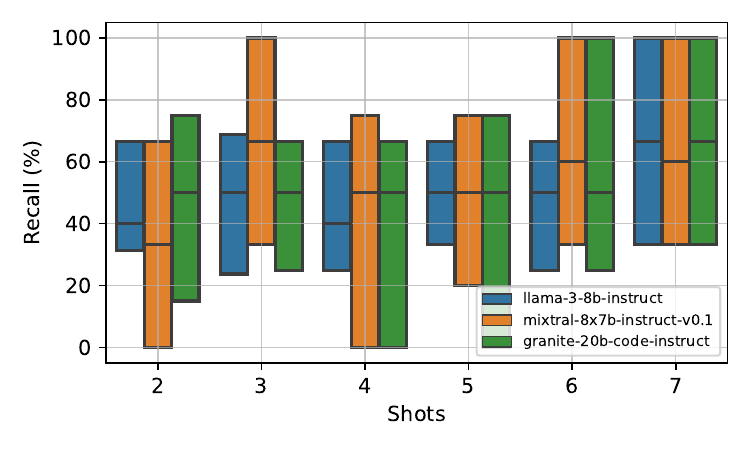}
    \caption{Variation in API recall for different number of in-context examples for ToolBench dataset}
    \label{fig:model:shots:variance}
\end{figure}

\paragraph{\noindent
\textbf{Increasing model size or number of examples does not mitigate order sensitivity.}}
We randomly sample 100 test instances from the ToolBench dataset \cite{lu2022toolbench}.
We vary the number of in-context examples between $2-7$, and use LLMs of varying sizes. We observe that adding more examples does not mitigate the prompt sensitivity, as illustrated in Figure~\ref{fig:model:shots:variance}.

\begin{figure}[htp]
    \centering
    \begin{subfigure}[t]{0.48\linewidth}
        \centering
        \includegraphics[width=0.95\linewidth]{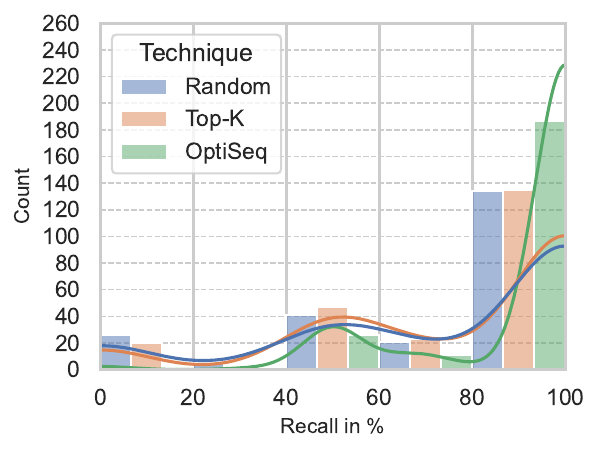}
        \caption{ToolBench dataset on \texttt{llama-3-8b-instruct}.}
        \label{fig:distribution:kde:tb:meta8b:1}
    \end{subfigure}
    \begin{subfigure}[t]{0.48\linewidth}
        \centering
        \includegraphics[width=0.95\linewidth]{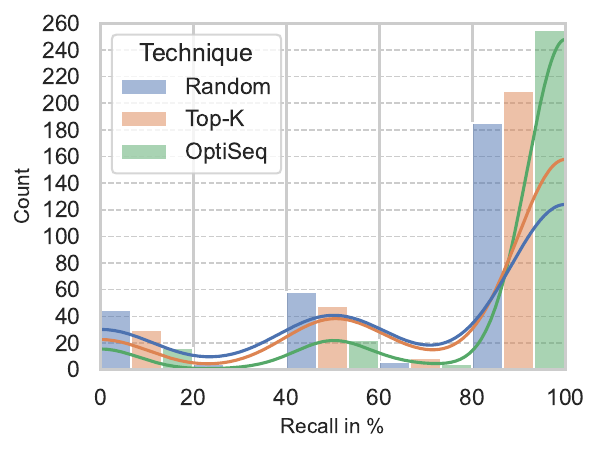}
        \caption{ToolBench dataset on \texttt{granite-20b-code-instruct}.}
        \label{fig:distribution:kde:tb:ibm20b:1}
    \end{subfigure}
    \begin{subfigure}[t]{0.48\linewidth}
        \centering
        \includegraphics[width=0.95\linewidth]{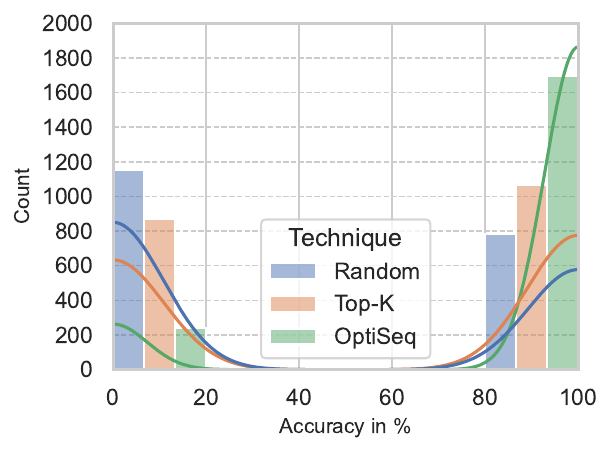}
        \caption{AG News dataset on \texttt{llama-3-8b-instruct}.}
        \label{fig:distribution:kde:agnews:meta8b}
    \end{subfigure}
    \begin{subfigure}[t]{0.48\linewidth}
        \centering
        \caption{AG News dataset on \texttt{granite-20b-code-instruct}.}
        \label{fig:distribution:kde:agnews:ibm20b}
    \end{subfigure}
    
    \caption{Distribution of Recall values for ToolBench (top row) and AG News (bottom row) datasets.}
    \label{fig:distribution:kde:combined:all}
\end{figure}
\textbf{\ours~improves the number of correct predictions as compared to Random and Top-K selection}. We sample test cases for different tasks and observe the performance spread. Figure~\ref{fig:distribution:kde:combined:all} shows the distribution of Recall values for API Sequence generation task and Accuracy values for the classification task. \ours~performs better than Random and Top-K selection -- shifts further towards the 100\% -- by increasing the number of correctly predicted sequences.

\begin{figure}
    \centering
    \includegraphics[width=0.8\linewidth]{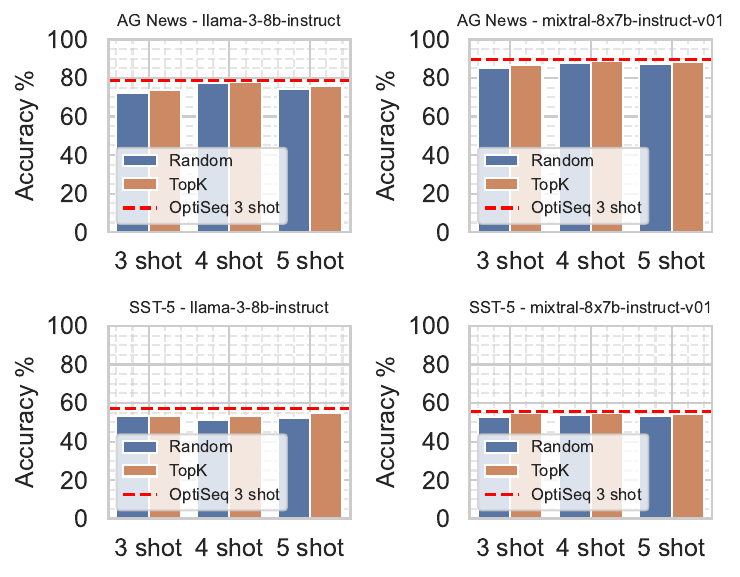}
    \caption{Comparing 3, 4, and 5 shot random/Top-K selection strategy with 3 shot \ours\ (in red) for the classification task.}
    \label{fig:345:shot:optiseq:class}
\end{figure}

Figure~\ref{fig:345:shot:optiseq:class} illustrates the performance of \ours\ on classification tasks, comparing it to 3-, 4-, and 5-shot in-context learning (ICL) using both random order and Top-K selection of examples. The results demonstrate that \ours, by strategically ordering a smaller set of examples, outperforms approaches using larger sets of examples in random or Top-K order.

\section{Appendix: Dataset Labels}

Table~\ref{table:labels:number} presents the number of possible labels for each dataset in our study. This information is crucial for understanding the complexity of the classification tasks the model must perform.

\begin{table}[]
\centering
\begin{tabular}{c|c|c}
\hline
\textbf{Dataset} & \textbf{Labels} & \textbf{Type} \\ \hline
ToolBench        & 100 & Multi-class            \\
RestGPT          & 75 & Multi-class             \\
AG\_news         & 4 & Single-class              \\
SST-5            & 5 & Single-class               \\
TREC             & 6 & Single-class             
\end{tabular}
\caption{Number of labels for each dataset}
\label{table:labels:number}
\end{table}

As shown in the table, ToolBench and RestGPT are multi-class classification tasks with 100 and 75 possible labels, respectively. These datasets present more complex classification challenges due to their higher number of potential outcomes. In contrast, AG\_news, SST-5, and TREC are single-class classification tasks with fewer labels (4, 5, and 6, respectively), representing comparatively simpler classification problems.

The variation in the number of labels and classification types across these datasets allows for a comprehensive evaluation of our model's performance across different levels of task complexity.

\section{Appendix: Prompt structure and sample inputs and outputs}

We present the prompt structures and sample inputs and outputs for each dataset we experimented with.

\include{toolbench_appendix}

\include{rest_gpt_appendix}

\include{ag_news_appendix}

\include{sst5_appendix}

\include{trec_appendix}

%% file: toolbench_appendix.tex
\twocolumn[
\subsection{ToolBench Dataset}
\vspace{-0.5em} 
The ToolBench dataset~\citep{lu2022toolbench} is a specialized benchmark designed to evaluate large language models (LLMs) on their ability to predict and sequence API calls in multi-tool reasoning tasks. It includes diverse natural language instructions paired with ground-truth API sequences across various domains and tools. Each data point typically contains an instruction, a catalog of available APIs, a set of in-context examples, and an evaluation metric (e.g., accuracy, precision, and recall). The dataset is tailored to assess the impact of in-context learning, permutation ordering, and tool usage alignment with human-designed workflows. ToolBench supports research on improving multi-tool coordination, mitigating biases in tool selection, and optimizing task-specific API predictions. For the ToolBench dataset in particular, we evaluate  with \textit{G1 – single-tool}, \textit{G2 – intra-category} multi-tool, and \textit{G3 – intra-collection multi-tool} parts of the dataset~\citep{lu2022toolbench}. We report aggregated results for all three parts of ToolBench to provide a concise summary. This approach enhances clarity, improves statistical robustness, and demonstrates the model's ability to generalize across tasks and scenarios. The prompt structure can be seen in~\ref{tab:toolbench:prompt}. We can see instances of in-context ordering sensitivity and identification of the relevant sequences using log probs in \ours\ in~\ref{tab:toolbench:instances}.
\vspace{1em} 
]
\begin{table*}[ht]
    \centering
    \footnotesize  
    \begin{tabular}{c}
        \hline 
        \\
        \begin{minipage}{\textwidth} 
            \begin{itemize}
                \item \textbf{Task Description:}
                \begin{itemize}
                    \item "I will ask you to perform a task. Your job is to come up with a sequence of APIs in a comma-separated list in the format that will perform the task. Start the list with << and end it with >>. Do not include anything other than the API name. Use the APIs below to answer the question posed to you. Avoid the use of any other text unless specified."
                \end{itemize}
                
                \item \textbf{APIs Available:}
                \begin{itemize}
                    \item \texttt{asin\_data.category}: Retrieve category results from Amazon.
                    \item \texttt{asin\_data.offers}: Retrieve seller offers for a product on Amazon.
                    \item \texttt{asin\_data.reviews}: Retrieve customer reviews for a product on Amazon.
                    \item \texttt{asin\_data.search}: Retrieve search results for an Amazon domain.
                    \item \texttt{asin\_data.product}: Retrieve details of a single product on Amazon.
                    \item \texttt{keyword\_analysis.popularsitesforquery}: Get popular sites for a search query.
                    \item \texttt{keyword\_analysis.similarqueries}: Get similar queries for a search query.
                    \item \texttt{spellout.rulesets}: List available rule sets for a given language.
                    \item \texttt{immersiverouletteapi.statistics}: Get statistics of wheel results.
                    \item \texttt{diffbot.article\_api}: Extract clean article text from web pages.
                    \item \texttt{covid\_19\_india.get\_details}: Get coronavirus updates for India.
                    \item \texttt{realtor\_data\_api\_for\_real\_estate.realtorpropertylist}: Get Realtor Property List.
                    \item \texttt{generate\_linkedin\_leads.get\_available\_locations}: Get available locations for LinkedIn leads.
                    \item \texttt{virtual\_number.get\_all\_countries}: Get list of available countries.
                    \item ...
                \end{itemize}

                \item \textbf{Examples:}
                \begin{itemize}
                    \item \textbf{Utterance:} "I am a fitness enthusiast and I want to buy a fitness tracker. Can you suggest some top-rated fitness trackers available on Amazon along with their features and prices?"
                    \item \textbf{Sequence:} \texttt{<<asin\_data.search, asin\_data.product>>}
                    
                    \item \textbf{Utterance:} "I'm a football enthusiast and I want to know more about Lionel Messi's career. Can you provide me with information about Messi's clubs, managers, teammates, and referees?"
                    \item \textbf{Sequence:} \texttt{<<theclique.transfermarkt\_search, theclique.transfermarkt\_details>>}
                    
                    \item \textbf{Utterance:} "I want to plan a surprise birthday party for my friend. Can you suggest popular sites and main keywords for the search query 'birthday party ideas'?"
                    \item \textbf{Sequence:} \texttt{<<keyword\_analysis.popularsitesforquery, keyword\_analysis.querykeywords>>}
                \end{itemize}
                \item \textbf{Test Utterance:}
                \begin{itemize}
                    \item \textbf{Utterance:} "I want to explore trending content on social media. Can you provide me with the current trending feed of videos? I would like to limit the output to 20 records. Please include the direct URLs to the videos and their statistics. Additionally, if possible, I would like to filter the feed based on a specific hashtag, such as \#summer."
                    \item \textbf{Sequence:}
                \end{itemize}
            \end{itemize}
        \end{minipage} \\
        \\
        \hline 
    \end{tabular}
    \normalsize 
    \caption{Prompt Structure for API Sequencing Task using ToolBench}
    \label{tab:toolbench:prompt}
\end{table*}

\begin{table*}[ht]
    \centering
    \small
    \resizebox{\textwidth}{!}{%

\begin{tabular}{p{3.5cm} p{5.5cm} c c c c c}
\toprule
\textbf{Utterance} & \textbf{Generated Sequence} & \textbf{Order} & \textbf{Precision} & \textbf{Recall} & \textbf{Accuracy} & \textbf{Log Prob \ours} \\
\midrule
\multirow{6}{*}{\parbox{3.5cm}{\tiny Utterance: I need to retrieve the pending messages from my device with ID 123456. Please provide the pending messages using my TrumpetBox Cloud API KEY.}} 
& \tiny \textcolor{blue}{\texttt{trumpetbox\_cloud.devices\_getasingledeviceinfofromaccount, trumpetbox\_cloud.messages\_getpendingmessagesfromaccount}} & 1  & \textbf{100.0} & \textbf{100.0} & \textbf{100.0} & \textbf{-1.54} \\
& \tiny \texttt{trumpetbox\_cloud.messages\_getpendingmessagesfromaccount} & 2  & 100.0 & 50.0 & 0.0 & -3.55 \\
& \tiny \textcolor{blue}{\texttt{trumpetbox\_cloud.devices\_getasingledeviceinfofromaccount, trumpetbox\_cloud.messages\_getpendingmessagesfromaccount}} & 3  & \textbf{100.0} & \textbf{100.0} & \textbf{100.0} & \textbf{-1.54} \\
& \tiny \texttt{trumpetbox\_cloud.messages\_getpendingmessagesfromaccount} & 4  & 100.0 & 50.0 & 0.0 & -3.55 \\
& \tiny \texttt{trumpetbox\_cloud.messages\_getpendingmessagesfromaccount} & 5  & 100.0 & 50.0 & 0.0 & -3.55 \\
& \tiny \textcolor{blue}{\texttt{trumpetbox\_cloud.devices\_getasingledeviceinfofromaccount, trumpetbox\_cloud.messages\_getpendingmessagesfromaccount}} & 6  & \textbf{100.0} & \textbf{100.0} & \textbf{100.0} & \textbf{-1.54} \\
\midrule
\multirow{6}{*}{\parbox{3.5cm}{\tiny Utterance: I need to track the performance of my family's crypto investments. Can you provide me with a list of our current positions for Bitcoin, Ethereum, and Ripple, along with their historical data and market trends?}} 
& \tiny \textcolor{blue}{\texttt{crypto\_grana.list.position.for.each.crypto, crypto\_grana.list.histories}} & 1  & \textbf{100.0} & \textbf{100.0} & \textbf{100.0} & \textbf{-1.67} \\
& \tiny \texttt{crypto\_grana.list.position.for.each.crypto} & 2  & 100.0 & 50.0 & 0.0 & -5.61 \\
& \tiny \texttt{crypto\_grana.list.position.for.each.crypto} & 3  & 100.0 & 50.0 & 0.0 & -5.61 \\
& \tiny \textcolor{blue}{\texttt{crypto\_grana.list.position.for.each.crypto, crypto\_grana.list.histories}} & 4  & \textbf{100.0} & \textbf{100.0} & \textbf{100.0} & \textbf{-1.67} \\
& \tiny \texttt{crypto\_grana.list.position.for.each.crypto} & 5  & 100.0 & 50.0 & 0.0 & -5.61 \\
& \tiny \textcolor{blue}{\texttt{crypto\_grana.list.position.for.each.crypto, crypto\_grana.list.histories}} & 6  & \textbf{100.0} & \textbf{100.0} & \textbf{100.0} & \textbf{-1.67} \\
\midrule
\multirow{6}{*}{\parbox{3.5cm}{\tiny Utterance: I run a company that organizes 4D lottery events. Can you give me the list of available 4D companies? It would also be helpful to get the past results from January 1, 2020, to March 31, 2020. Finally, I need the 4D results for Magnum on January 29, 2020.}} 
& \tiny \textcolor{blue}{\texttt{4d\_results.get.4d.company.list, 4d\_results.get.past.results.(1.year), 4d\_results.get.4d.results}} & 1  & \textbf{100.0} & \textbf{100.0} & \textbf{100.0} & \textbf{-3.49} \\
& \tiny \textcolor{blue}{\texttt{4d\_results.get.4d.company.list, 4d\_results.get.past.results.(1.year), 4d\_results.get.4d.results}} & 2  & \textbf{100.0} & \textbf{100.0} & \textbf{100.0} & \textbf{-3.49} \\
& \tiny \textcolor{blue}{\texttt{4d\_results.get.4d.company.list, 4d\_results.get.past.results.(1.year), 4d\_results.get.4d.results}} & 3  & \textbf{100.0} & \textbf{100.0} & \textbf{100.0} & \textbf{-3.49} \\
& \tiny \textcolor{blue}{\texttt{4d\_results.get.4d.company.list, 4d\_results.get.past.results.(1.year), 4d\_results.get.4d.results}} & 4  & \textbf{100.0} & \textbf{100.0} & \textbf{100.0} & \textbf{-3.49} \\
& \tiny \texttt{4d\_results.get.4d.company.list, 4d\_results.get.past.results.(1.year)} & 5  & 100.0 & 66.67 & 0.0 & -4.55 \\
& \tiny \textcolor{blue}{\texttt{4d\_results.get.4d.company.list, 4d\_results.get.past.results.(1.year), 4d\_results.get.4d.results}} & 6  & \textbf{100.0} & \textbf{100.0} & \textbf{100.0} & \textbf{-3.49} \\
\midrule
\multirow{6}{*}{\parbox{3.5cm}{\tiny Utterance: I'm a weather enthusiast and I'm interested in studying aviation weather data. Can you provide me with the most recent TAFs for the next 24 hours? I also want to see the most recent METARs from the past 2 hours. Please include the temperature, dew point, and wind direction in both reports.}} 
& \tiny \texttt{aviation\_weather\_center.most\_recent.tafs, aviation\_weather\_center.most\_recent.metars} & 1  & 0.0 & 0.0 & 0.0 & -6.18 \\
& \tiny \textcolor{blue}{\texttt{aviation\_weather\_center.most.recent.tafs, aviation\_weather\_center.most.recent.metars}} & 2  & \textbf{100.0} & \textbf{100.0} & \textbf{100.0} & \textbf{-4.17} \\
& \tiny \textcolor{blue}{\texttt{aviation\_weather\_center.most.recent.tafs, aviation\_weather\_center.most.recent.metars}} & 3  & \textbf{100.0} & \textbf{100.0} & \textbf{100.0} & \textbf{-4.17} \\
& \tiny \textcolor{blue}{\texttt{aviation\_weather\_center.most.recent.tafs, aviation\_weather\_center.most.recent.metars}} & 4  & \textbf{100.0} & \textbf{100.0} & \textbf{100.0} & \textbf{-4.17} \\
& \tiny \textcolor{blue}{\texttt{aviation\_weather\_center.most.recent.tafs, aviation\_weather\_center.most.recent.metars}} & 5  & \textbf{100.0} & \textbf{100.0} & \textbf{100.0} & \textbf{-4.17} \\
& \tiny \texttt{aviation\_weather\_center.most\_recent.tafs, aviation\_weather\_center.most\_recent.metars} & 6  & 0.0 & 0.0 & 0.0 & -6.18 \\
\midrule
\multirow{6}{*}{\parbox{3.5cm}{\tiny Utterance: I'm developing an art events app and I need a list of all genres of the events. Can you provide me with this information? It would be helpful if you could also give me a list of all locations where art events take place.}} 
& \tiny \textcolor{blue}{\texttt{art\_openings\_italy.get.all.genres, art\_openings\_italy.get.all.locations}} & 1  & \textbf{100.0} & \textbf{100.0} & \textbf{100.0} & \textbf{-1.12} \\
& \tiny \textcolor{blue}{\texttt{art\_openings\_italy.get.all.genres, art\_openings\_italy.get.all.locations}} & 2  & \textbf{100.0} & \textbf{100.0} & \textbf{100.0} & \textbf{-1.12} \\
& \tiny \textcolor{blue}{\texttt{art\_openings\_italy.get.all.genres, art\_openings\_italy.get.all.locations}} & 3  & \textbf{100.0} & \textbf{100.0} & \textbf{100.0} & \textbf{-1.12} \\
& \tiny \textcolor{blue}{\texttt{art\_openings\_italy.get.all.genres, art\_openings\_italy.get.all.locations}} & 4  & \textbf{100.0} & \textbf{100.0} & \textbf{100.0} & \textbf{-1.12} \\
& \tiny \textcolor{blue}{\texttt{art\_openings\_italy.get.all.genres, art\_openings\_italy.get.all.locations}} & 5  & \textbf{100.0} & \textbf{100.0} & \textbf{100.0} & \textbf{-1.12} \\
& \tiny \texttt{art\_openings\_italy.get\_all.genres, art\_openings\_italy.get\_all.locations} & 6  & 0.0 & 0.0 & 0.0 & -3.85 \\
\midrule
\multirow{6}{*}{\parbox{3.5cm}{\tiny Utterance: I'm planning a family vacation and need to check the availability of a specific product. Can you provide me with the details of the product 'XYZ' including its price, stock quantity, and ID? Additionally, I want to find a client with the name 'Mark' and his contact details.}} 
& \tiny \texttt{realtor\_data\_api\_for\_real\_estate.realtorpropertylist, realtor\_data\_api\_for\_real\_estate.realtoragentlist} & 1  & 0.0 & 0.0 & 0.0 & -8.30 \\
& \tiny \textcolor{blue}{\texttt{capacitacionangular.productos, capacitacionangular.cliente}} & 2  & \textbf{100.0} & \textbf{100.0} & \textbf{100.0} & \textbf{-6.69} \\
& \tiny \texttt{realtor\_data\_api\_for\_real\_estate.realtorpropertylist, realtor\_data\_api\_for\_real\_estate.realtoragentlist} & 3  & 0.0 & 0.0 & 0.0 & -8.30 \\
& \tiny \textcolor{blue}{\texttt{capacitacionangular.productos, capacitacionangular.cliente}} & 4  & \textbf{100.0} & \textbf{100.0} & \textbf{100.0} & \textbf{-6.69} \\
& \tiny \textcolor{blue}{\texttt{capacitacionangular.productos, capacitacionangular.cliente}} & 5  & \textbf{100.0} & \textbf{100.0} & \textbf{100.0} & \textbf{-6.69} \\
& \tiny \texttt{realtor\_data\_api\_for\_real\_estate.realtorpropertylist, realtor\_data\_api\_for\_real\_estate.realtoragentlist} & 6  & 0.0 & 0.0 & 0.0 & -8.30 \\
\midrule
\multirow{6}{*}{\parbox{3.5cm}{\tiny Utterance: I'm planning a vacation to Bali and I want to explore the most popular Instagram profiles from the area. Can you provide me with the Instagram profiles of famous travelers who have visited Bali recently? Additionally, I would like to see their stories.}} 
& \tiny \textcolor{blue}{\texttt{access\_instagram.instagram.endpoint\_copy, access\_instagram.instagram.endpoint}} & 1  & \textbf{100.0} & \textbf{100.0} & \textbf{100.0} & \textbf{-3.09} \\
& \tiny \texttt{access\_instagram.instagram.endpoint\_copy} & 2  & 100.0 & 50.0 & 0.0 & -3.39 \\
& \tiny \texttt{access\_instagram.instagram.endpoint\_copy} & 3  & 100.0 & 50.0 & 0.0 & -3.39 \\
& \tiny \textcolor{blue}{\texttt{access\_instagram.instagram.endpoint\_copy, access\_instagram.instagram.endpoint}} & 4  & \textbf{100.0} & \textbf{100.0} & \textbf{100.0} & \textbf{-3.09} \\
& \tiny \textcolor{blue}{\texttt{access\_instagram.instagram.endpoint\_copy, access\_instagram.instagram.endpoint}} & 5  & \textbf{100.0} & \textbf{100.0} & \textbf{100.0} & \textbf{-3.09} \\
& \tiny \textcolor{blue}{\texttt{access\_instagram.instagram.endpoint\_copy, access\_instagram.instagram.endpoint}} & 6  & \textbf{100.0} & \textbf{100.0} & \textbf{100.0} & \textbf{-3.09} \\
\midrule
\bottomrule
\end{tabular}%

    }
    \caption{Generated Sequences and Metrics for the ToolBench Dataset with \texttt{meta/llama-3-8b-instruct}}
    \label{tab:toolbench:instances}
\end{table*}

%% file: rest_gpt_appendix.tex
\twocolumn[
\subsection{RestGPT Dataset}
\vspace{-0.5em} 
The RestBench~\citep{wu2023restgpt} dataset is a high-quality test set designed to evaluate large language models (LLMs) on task execution in two primary domains: the TMDB movie database and the Spotify music player. It includes natural language queries and instructions that require models to reason about and generate API call sequences for tasks such as retrieving movie details, searching for music tracks, creating playlists, and handling user preferences. Each data point comprises a user query, a structured API catalog, and ground-truth API sequences, with a focus on multi-step reasoning and alignment with user intents. RestBench serves as a robust benchmark for assessing the capabilities of LLMs in handling complex domain-specific workflows, demonstrating their potential in real-world applications across entertainment platforms.
 The prompt structure can be seen in~\ref{tab:restgpt:prompt}. We can see instances of in-context ordering sensitivity and identification of the relevant sequences using log probs in \ours\ in~\ref{tab:restgpt:instances}.
\vspace{1em} 
]

\begin{table*}[h]
    \centering
    \footnotesize 
    \begin{tabular}{c}
        \hline 
        \\
        \begin{minipage}{\textwidth} 
            \begin{itemize}
                \item \textbf{Task Description:}
                \begin{itemize}
                    \item "I will ask you to perform a task. Your job is to come up with a sequence of APIs in a comma-separated list in the format that will perform the task. Start the list with << and end it with >>. Do not include anything other than the API name. Use the APIs below to answer the question posed to you. Avoid the use of any other text unless specified."
                \end{itemize}
                
                \item \textbf{APIs Available:}
                \begin{itemize}
                    \item \texttt{SearchCollection}: GET /search/collection
                    \item \texttt{CollectionImages}: GET /collection/{collection\_id}/images
                    \item \texttt{SearchPerson}: GET /search/person
                    \item \texttt{PersonMovieCredits}: GET /person/{person\_id}/movie\_credits
                    \item \texttt{SearchMovie}: GET /search/movie
                    \item \texttt{MovieCredits}: GET /movie/{movie\_id}/credits
                    \item \texttt{MovieTopRated}: GET /movie/top\_rated
                    \item \texttt{SearchCompany}: GET /search/company
                    \item \texttt{CompanyImages}: GET /company/{company\_id}/images
                    \item \texttt{PersonImages}: GET /person/{person\_id}/images
                    \item \texttt{MovieSimilar}: GET /movie/{movie\_id}/similar
                    \item \texttt{MovieReviews}: GET /movie/{movie\_id}/reviews
                    \item \texttt{MovieRecommendations}: GET /movie/{movie\_id}/recommendations
                    \item \texttt{PersonTvCredits}: GET /person/{person\_id}/tv\_credits
                    \item \texttt{SearchTv}: GET /search/tv
                    \item \texttt{TvRecommendations}: GET /tv/{tv\_id}/recommendations
                    \item \texttt{Trending}: GET /trending/{media\_type}/{time\_window}
                    \item \texttt{.....}
                \end{itemize}

                \item \textbf{Examples:}
                \begin{itemize}
                    \item \textbf{Utterance:} "Give me the number of movies directed by Sofia Coppola."
                    \item \textbf{Sequence:} \texttt{<<SearchPerson, PersonMovieCredits>>}
                    
                    \item \textbf{Utterance:} "Who was the lead actor in the movie The Dark Knight?"
                    \item \textbf{Sequence:} \texttt{<<SearchMovie, MovieCredits>>}
                    
                    \item \textbf{Utterance:} "Who directed the top-1 rated movie?"
                    \item \textbf{Sequence:} \texttt{<<MovieTopRated, MovieCredits>>}

                \end{itemize}
                \item \textbf{Test Utterance:}
                \begin{itemize}
                    \item \textbf{Utterance:} "I'm watching the tv series The Last Of Us and I need some more recommendations"
                    \item \textbf{Sequence:} \texttt{}
                \end{itemize}
            \end{itemize}
        \end{minipage} \\
        \\
        \hline 
    \end{tabular}
    \normalsize 
    \caption{Prompt Structure for API Sequencing Tasks in RestGPT}
    \label{tab:restgpt:prompt}
\end{table*}

\begin{table*}[ht]
    \centering
    \small
    \resizebox{\textwidth}{!}{%

\begin{tabular}{p{3.5cm} p{5.5cm} c c c c c}
\toprule
\textbf{Utterance} & \textbf{Generated Sequence} & \textbf{Order} & \textbf{Precision} & \textbf{Recall} & \textbf{Accuracy} & \textbf{Log Prob \ours} \\
\midrule
\multirow{6}{*}{\parbox{3.5cm}{\tiny Utterance: I just finished watching Titanic and I want some other movie recommendations}} 
& \tiny \texttt{Movie, MovieRecommendations} & 1  & 50.0 & 50.0 & 0.0 & -18.28 \\
& \tiny \textcolor{blue}{\texttt{SearchMovie, MovieRecommendations}} & 2  & \textbf{100.0} & \textbf{100.0} & \textbf{100.0} & \textbf{-6.76} \\
& \tiny \texttt{MovieRecommendations} & 3  & 100.0 & 50.0 & 0.0 & -9.56 \\
& \tiny \texttt{Movie, MovieRecommendations} & 4  & 50.0 & 50.0 & 0.0 & -19.59 \\
& \tiny \texttt{Movie, MovieRecommendations} & 5  & 50.0 & 50.0 & 0.0 & -18.79 \\
& \tiny \texttt{Movie, MovieRecommendations} & 6  & 50.0 & 50.0 & 0.0 & -19.14 \\
\midrule
\multirow{6}{*}{\parbox{3.5cm}{\tiny Utterance: What dose the lead actor of Titanic look like?}} 
& \tiny \textcolor{blue}{\texttt{SearchMovie, MovieCredits, PersonImages}} & 1  & \textbf{100.0} & \textbf{100.0} & \textbf{100.0} & \textbf{-5.48} \\
& \tiny \textcolor{blue}{\texttt{SearchMovie, MovieCredits, PersonImages}} & 2  & \textbf{100.0} & \textbf{100.0} & \textbf{100.0} & \textbf{-6.62} \\
& \tiny \textcolor{blue}{\texttt{SearchMovie, MovieCredits, PersonImages}} & 3  & \textbf{100.0} & \textbf{100.0} & \textbf{100.0} & \textbf{-7.25} \\
& \tiny \texttt{SearchPerson, PersonImages} & 4  & 50.0 & 33.33 & 0.0 & -16.95 \\
& \tiny \textcolor{blue}{\texttt{SearchMovie, MovieCredits, PersonImages}} & 5  & \textbf{100.0} & \textbf{100.0} & \textbf{100.0} & \textbf{-5.83} \\
& \tiny \texttt{SearchPerson, PersonImages} & 6  & 50.0 & 33.33 & 0.0 & -17.07 \\
\midrule
\multirow{6}{*}{\parbox{3.5cm}{\tiny Utterance: What is the logo of the Walt Disney?}} 
& \tiny \textcolor{blue}{\texttt{SearchCompany, CompanyImages}} & 1  & \textbf{100.0} & \textbf{100.0} & \textbf{100.0} & \textbf{-5.09} \\
& \tiny \textcolor{blue}{\texttt{SearchCompany, CompanyImages}} & 2  & \textbf{100.0} & \textbf{100.0} & \textbf{100.0} & \textbf{-5.83} \\
& \tiny \textcolor{blue}{\texttt{SearchCompany, CompanyImages}} & 3  & \textbf{100.0} & \textbf{100.0} & \textbf{100.0} & \textbf{-5.69} \\
& \tiny \textcolor{blue}{\texttt{SearchCompany, CompanyImages}} & 4  & \textbf{100.0} & \textbf{100.0} & \textbf{100.0} & \textbf{-5.32} \\
& \tiny \texttt{SearchCompany, CompanyImages, Movie, MovieReleaseDates, andendwitha, SearchMovie, Movie} & 5  & 28.57 & 100.0 & 0.0 & -94.40 \\
& \tiny \textcolor{blue}{\texttt{SearchCompany, CompanyImages}} & 6  & \textbf{100.0} & \textbf{100.0} & \textbf{100.0} & \textbf{-5.32} \\
\midrule
\multirow{6}{*}{\parbox{3.5cm}{\tiny Utterance: Who directed the top-1 rated movie?}} 
& \tiny \texttt{MovieTopRated, Movie, Credit} & 1  & 33.33 & 50.0 & 0.0 & -35.84 \\
& \tiny \textcolor{blue}{\texttt{MovieTopRated, MovieCredits}} & 2  & \textbf{100.0} & \textbf{100.0} & \textbf{100.0} & \textbf{-6.34} \\
& \tiny \textcolor{blue}{\texttt{MovieTopRated, MovieCredits}} & 3  & \textbf{100.0} & \textbf{100.0} & \textbf{100.0} & \textbf{-6.89} \\
& \tiny \textcolor{blue}{\texttt{MovieTopRated, MovieCredits}} & 4  & \textbf{100.0} & \textbf{100.0} & \textbf{100.0} & \textbf{-6.62} \\
& \tiny \textcolor{blue}{\texttt{MovieTopRated, MovieCredits}} & 5  & \textbf{100.0} & \textbf{100.0} & \textbf{100.0} & \textbf{-6.95} \\
& \tiny \textcolor{blue}{\texttt{MovieTopRated, MovieCredits}} & 6  & \textbf{100.0} & \textbf{100.0} & \textbf{100.0} & \textbf{-6.71} \\
\midrule
\multirow{6}{*}{\parbox{3.5cm}{\tiny Utterance: Who is the director of the movie "Twilight"?}} 
& \tiny \textcolor{blue}{\texttt{SearchMovie, MovieCredits}} & 1  & \textbf{100.0} & \textbf{100.0} & \textbf{100.0} & \textbf{-5.71} \\
& \tiny \texttt{SearchMovie, Mcosy.credits} & 2  & 50.0 & 50.0 & 0.0 & -54.40 \\
& \tiny \textcolor{blue}{\texttt{SearchMovie, MovieCredits}} & 3  & \textbf{100.0} & \textbf{100.0} & \textbf{100.0} & \textbf{-7.08} \\
& \tiny \textcolor{blue}{\texttt{SearchMovie, MovieCredits}} & 4  & \textbf{100.0} & \textbf{100.0} & \textbf{100.0} & \textbf{-6.59} \\
& \tiny \textcolor{blue}{\texttt{SearchMovie, MovieCredits}} & 5  & \textbf{100.0} & \textbf{100.0} & \textbf{100.0} & \textbf{-5.69} \\
& \tiny \textcolor{blue}{\texttt{SearchMovie, MovieCredits}} & 6  & \textbf{100.0} & \textbf{100.0} & \textbf{100.0} & \textbf{-5.83} \\
\midrule
\multirow{6}{*}{\parbox{3.5cm}{\tiny Utterance: Who is the most popular person?}} 
& \tiny \textcolor{blue}{\texttt{PersonPopular}} & 1  & \textbf{100.0} & \textbf{100.0} & \textbf{100.0} & \textbf{-0.22} \\
& \tiny \texttt{Trending, PersonPopular} & 2  & 50.0 & 100.0 & 0.0 & -16.97 \\
& \tiny \textcolor{blue}{\texttt{PersonPopular}} & 3  & \textbf{100.0} & \textbf{100.0} & \textbf{100.0} & \textbf{-0.10} \\
& \tiny \textcolor{blue}{\texttt{PersonPopular}} & 4  & \textbf{100.0} & \textbf{100.0} & \textbf{100.0} & \textbf{-0.10} \\
& \tiny \textcolor{blue}{\texttt{PersonPopular}} & 5  & \textbf{100.0} & \textbf{100.0} & \textbf{100.0} & \textbf{-0.80} \\
& \tiny \textcolor{blue}{\texttt{PersonPopular}} & 6  & \textbf{100.0} & \textbf{100.0} & \textbf{100.0} & \textbf{-0.10} \\
\midrule
\multirow{6}{*}{\parbox{3.5cm}{\tiny Utterance: Who was the lead actor in the movie The Dark Knight?}} 
& \tiny \textcolor{blue}{\texttt{SearchMovie, MovieCredits}} & 1  & \textbf{100.0} & \textbf{100.0} & \textbf{100.0} & \textbf{-6.86} \\
& \tiny \textcolor{blue}{\texttt{SearchMovie, MovieCredits}} & 2  & \textbf{100.0} & \textbf{100.0} & \textbf{100.0} & \textbf{-7.35} \\
& \tiny \texttt{SearchMovie, MovieCredits, ortrailing} & 3  & 66.67 & 100.0 & 0.0 & -54.92 \\
& \tiny \textcolor{blue}{\texttt{SearchMovie, MovieCredits}} & 4  & \textbf{100.0} & \textbf{100.0} & \textbf{100.0} & \textbf{-7.23} \\
& \tiny \textcolor{blue}{\texttt{SearchMovie, MovieCredits}} & 5  & \textbf{100.0} & \textbf{100.0} & \textbf{100.0} & \textbf{-7.22} \\
& \tiny \texttt{SearchMovie, B} & 6  & 50.0 & 50.0 & 0.0 & -36.12 \\
\midrule
\multirow{6}{*}{\parbox{3.5cm}{\tiny Utterance: give me a image for the collection Star Wars}} 
& \tiny \textcolor{blue}{\texttt{SearchCollection, CollectionImages}} & 1  & \textbf{100.0} & \textbf{100.0} & \textbf{100.0} & \textbf{-5.37} \\
& \tiny \textcolor{blue}{\texttt{SearchCollection, CollectionImages}} & 2  & \textbf{100.0} & \textbf{100.0} & \textbf{100.0} & \textbf{-5.69} \\
& \tiny \texttt{CollectionImages} & 3  & 100.0 & 50.0 & 0.0 & -5.75 \\
& \tiny \textcolor{blue}{\texttt{SearchCollection, CollectionImages}} & 4  & \textbf{100.0} & \textbf{100.0} & \textbf{100.0} & \textbf{-5.60} \\
& \tiny \textcolor{blue}{\texttt{SearchCollection, CollectionImages}} & 5  & \textbf{100.0} & \textbf{100.0} & \textbf{100.0} & \textbf{-5.55} \\
& \tiny \textcolor{blue}{\texttt{SearchCollection, CollectionImages}} & 6  & \textbf{100.0} & \textbf{100.0} & \textbf{100.0} & \textbf{-5.35} \\
\midrule
\multirow{6}{*}{\parbox{3.5cm}{\tiny Utterance: give me a photo belong to the second episode of the first season of the Witcher}} 
& \tiny \textcolor{blue}{\texttt{SearchTv, TvSeasonEpisodeImages}} & 1  & \textbf{100.0} & \textbf{100.0} & \textbf{100.0} & \textbf{-6.13} \\
& \tiny \textcolor{blue}{\texttt{SearchTv, TvSeasonEpisodeImages}} & 2  & \textbf{100.0} & \textbf{100.0} & \textbf{100.0} & \textbf{-6.36} \\
& \tiny \texttt{TvSeasonEpisodeImages, } & 3  & 50.0 & 50.0 & 0.0 & -20.16 \\
& \tiny \textcolor{blue}{\texttt{SearchTv, TvSeasonEpisodeImages}} & 4  & \textbf{100.0} & \textbf{100.0} & \textbf{100.0} & \textbf{-6.36} \\
& \tiny \textcolor{blue}{\texttt{SearchTv, TvSeasonEpisodeImages}} & 5  & \textbf{100.0} & \textbf{100.0} & \textbf{100.0} & \textbf{-6.60} \\
& \tiny \textcolor{blue}{\texttt{SearchTv, TvSeasonEpisodeImages}} & 6  & \textbf{100.0} & \textbf{100.0} & \textbf{100.0} & \textbf{-6.23} \\
\midrule
\multirow{6}{*}{\parbox{3.5cm}{\tiny Utterance: give me the number of movies directed by Sofia Coppola}} 
& \tiny \textcolor{blue}{\texttt{SearchPerson, PersonMovieCredits}} & 1  & \textbf{100.0} & \textbf{100.0} & \textbf{100.0} & \textbf{-6.60} \\
& \tiny \textcolor{blue}{\texttt{SearchPerson, PersonMovieCredits}} & 2  & \textbf{100.0} & \textbf{100.0} & \textbf{100.0} & \textbf{-6.35} \\
& \tiny \textcolor{blue}{\texttt{SearchPerson, PersonMovieCredits}} & 3  & \textbf{100.0} & \textbf{100.0} & \textbf{100.0} & \textbf{-7.14} \\
& \tiny \texttt{SearchPerson, PersonMovieCredits, andenditwitha} & 4  & 66.67 & 100.0 & 0.0 & -55.08 \\
& \tiny \textcolor{blue}{\texttt{SearchPerson, PersonMovieCredits}} & 5  & \textbf{100.0} & \textbf{100.0} & \textbf{100.0} & \textbf{-6.72} \\
& \tiny \texttt{SearchPerson, PersonMovieCredits, MOVIE\_TOP\_RATED} & 6  & 66.67 & 100.0 & 0.0 & -33.02 \\
\midrule
\multirow{6}{*}{\parbox{3.5cm}{\tiny Utterance: tell me a TV show recently directed by Catherine Hardwicke}} 
& \tiny \textcolor{blue}{\texttt{SearchPerson, PersonTvCredits}} & 1  & \textbf{100.0} & \textbf{100.0} & \textbf{100.0} & \textbf{-5.31} \\
& \tiny \texttt{SearchPerson, PersonTvCredits, Tv, andenditwitha} & 2  & 50.0 & 100.0 & 0.0 & -52.12 \\
& \tiny \texttt{SearchCompany, CompanyImages, SearchPerson, PersonTvCredits} & 3  & 50.0 & 100.0 & 0.0 & -25.36 \\
& \tiny \texttt{Trending, TvCredits, DetectcurrentTVshow} & 4  & 0.0 & 0.0 & 0.0 & -60.15 \\
& \tiny \textcolor{blue}{\texttt{SearchPerson, PersonTvCredits}} & 5  & \textbf{100.0} & \textbf{100.0} & \textbf{100.0} & \textbf{-5.31} \\
& \tiny \texttt{SearchPerson, PersonTvCredits, Trending, movie\_type=tv, time\_window=week, TvCredits} & 6  & 33.33 & 100.0 & 0.0 & -59.68 \\
\midrule
\bottomrule
\end{tabular}%

    }
    \caption{Generated Sequences and Metrics for the RestGPT Dataset with \texttt{meta/llama-3-8b-instruct}}
    \label{tab:restgpt:instances}
\end{table*}

%% file: ag_news_appendix.tex
\begin{table*}[h]
    \centering
    \footnotesize 
    \begin{tabular}{c}
        \hline 
        \\
        \begin{minipage}{\textwidth} 
            \begin{itemize}
                \item \textbf{Task Description:}
                \begin{itemize}
                    \item "Classify the following news articles into one of these categories: World, Sports, Business, Sci/Tech."
                \end{itemize}

                \item \textbf{Examples:}
                \begin{itemize}
                    \item \textbf{Title:} "Fears for T N pension after talks"
                    \item \textbf{Article:} "Unions representing workers at Turner Newall say they are 'disappointed' after talks with stricken parent firm Federal Mogul."
                    \item \textbf{Category:} \texttt{Business}

                    \item \textbf{Title:} "The Race is On: Second Private Team Sets Launch Date for Human Spaceflight (SPACE.com)"
                    \item \textbf{Article:} "SPACE.com - TORONTO, Canada -- A second team of rocketeers competing for the \$10 million Ansari X Prize, a contest for privately funded suborbital space flight, has officially announced the first launch date for its manned rocket."
                    \item \textbf{Category:} \texttt{Sci/Tech}

                    \item \textbf{Title:} "Giddy Phelps Touches Gold for First Time"
                    \item \textbf{Article:} "Michael Phelps won the gold medal in the 400 individual medley and set a world record in a time of 4 minutes 8.26 seconds."
                    \item \textbf{Category:} \texttt{Sports}
                    \end{itemize}
                    \item \textbf{Test Utterance}
                    \begin{itemize}
                    \item \textbf{Title:} "Prediction Unit Helps Forecast Wildfires (AP)"
                    \item \textbf{Article:} "AP - It's barely dawn when Mike Fitzpatrick starts his shift with a blur of colorful maps, figures and endless charts, but already he knows what the day will bring. Lightning will strike in places he expects. Winds will pick up, moist places will dry and flames will roar."
                    \item \textbf{Category:} \texttt{}
                \end{itemize}
            \end{itemize}
        \end{minipage} \\
        \\
        \hline 
    \end{tabular}
    \normalsize 
    \caption{AG News Prompt Structure}
    \label{tab:ag_news_prompt}
\end{table*}

\begin{table*}[ht]
    \centering
    \small
    \resizebox{\textwidth}{!}{%

\begin{tabular}{p{4.0cm} p{3.5cm} c c c }
\toprule
\textbf{Utterance} & \textbf{Generated Sequence} & \textbf{Order} & \textbf{Accuracy} & \textbf{Log Prob \ours} \\
\midrule
\multirow{6}{*}{\parbox{3.5cm}{\tiny Article: A  quot;formidable information and technology management challenge quot; faces the Homeland Security Department, according a report released today by the Government Accountability Office.}} 
& \texttt{World} & 1 & 0.0 & -1.57 \\
& \texttt{World} & 2 & 0.0 & -0.64 \\
& \texttt{World} & 3 & 0.0 & -0.77 \\
& \texttt{World} & 4 & 0.0 & -1.15 \\
& \textcolor{blue}{\texttt{Sci/Tech}} & 5 & \textbf{100.0} & \textbf{-0.55} \\
& \textcolor{blue}{\texttt{Sci/Tech}} & 6 & \textbf{100.0} & \textbf{-0.81} \\
\midrule
\multirow{6}{*}{\parbox{3.5cm}{\tiny Article: A former executive who was a participant in the wrongdoing that helped cripple Enron testified on Monday, providing the first glimpse through the eyes of a principal of}} 
& \textcolor{blue}{\texttt{Business}} & 1 & \textbf{100.0} & \textbf{-0.08} \\
& \texttt{World} & 2 & 0.0 & -1.34 \\
& \textcolor{blue}{\texttt{Business}} & 3 & \textbf{100.0} & \textbf{-0.16} \\
& \texttt{World} & 4 & 0.0 & -2.29 \\
& \texttt{World} & 5 & 0.0 & -0.87 \\
& \texttt{World} & 6 & 0.0 & -1.49 \\
\midrule
\multirow{6}{*}{\parbox{3.5cm}{\tiny Article: AOL has kicked off an initiative designed to make it easier for developers to engineer, test and distribute licensed AOL Instant Messenger (AIM) clients for mobile devices.}} 
& \textcolor{blue}{\texttt{Sci/Tech}} & 1 & \textbf{100.0} & \textbf{-0.29} \\
& \textcolor{blue}{\texttt{Sci/Tech}} & 2 & \textbf{100.0} & \textbf{-0.28} \\
& \textcolor{blue}{\texttt{Sci/Tech}} & 3 & \textbf{100.0} & \textbf{-0.28} \\
& \texttt{Business} & 4 & 0.0 & -0.59 \\
& \textcolor{blue}{\texttt{Sci/Tech}} & 5 & \textbf{100.0} & \textbf{-0.30} \\
& \texttt{Business} & 6 & 0.0 & -0.56 \\
\midrule
\multirow{6}{*}{\parbox{3.5cm}{\tiny Article: AP - Raymond Goethals, the Belgian soccer coach who led Olympique Marseille to the 1993 European Champions Cup title, died Monday, according to news reports. He was 83.}} 
& \textcolor{blue}{\texttt{Sports}} & 1 & \textbf{100.0} & \textbf{-0.17} \\
& \textcolor{blue}{\texttt{Sports}} & 2 & \textbf{100.0} & \textbf{-0.14} \\
& \texttt{World} & 3 & 0.0 & -1.07 \\
& \textcolor{blue}{\texttt{Sports}} & 4 & \textbf{100.0} & \textbf{-0.24} \\
& \textcolor{blue}{\texttt{Sports}} & 5 & \textbf{100.0} & \textbf{-0.04} \\
& \textcolor{blue}{\texttt{Sports}} & 6 & \textbf{100.0} & \textbf{-0.05} \\
\midrule
\midrule
\multirow{6}{*}{\parbox{3.5cm}{\tiny Article: Crude oil prices settled at \$49.64 a barrel, up 76 cents as traders expressed concern that recent hurricanes had hurt output in the United States.}} 
& \textcolor{blue}{\texttt{Business}} & 1 & \textbf{100.0} & \textbf{-0.14} \\
& \texttt{World} & 2 & 0.0 & -1.04 \\
& \texttt{World} & 3 & 0.0 & -1.92 \\
& \texttt{World} & 4 & 0.0 & -0.85 \\
& \texttt{World} & 5 & 0.0 & -0.89 \\
& \texttt{World} & 6 & 0.0 & -1.55 \\
\midrule
\multirow{6}{*}{\parbox{3.5cm}{\tiny Article: Moises Alou has a right to his opinion, Chicago Cubs manager Dusty Baker said Monday. Alou said everything he needed to say Sunday.}} 
& \textcolor{blue}{\texttt{Sports}} & 1 & \textbf{100.0} & \textbf{-0.04} \\
& \texttt{World} & 2 & 0.0 & -2.14 \\
& \textcolor{blue}{\texttt{Sports}} & 3 & \textbf{100.0} & \textbf{-0.08} \\
& \textcolor{blue}{\texttt{Sports}} & 4 & \textbf{100.0} & \textbf{-0.11} \\
& \textcolor{blue}{\texttt{Sports}} & 5 & \textbf{100.0} & \textbf{-0.02} \\
& \textcolor{blue}{\texttt{Sports}} & 6 & \textbf{100.0} & \textbf{-0.02} \\
\midrule
\multirow{6}{*}{\parbox{3.5cm}{\tiny Article: NEW YORK -- Dale Earnhardt Jr. has trouble remembering those frantic seconds when he escaped from his burning racecar. He believes, however, that his late father figured in his survival.}} 
& \textcolor{blue}{\texttt{Sports}} & 1 & \textbf{100.0} & \textbf{-0.10} \\
& \texttt{World} & 2 & 0.0 & -1.09 \\
& \textcolor{blue}{\texttt{Sports}} & 3 & \textbf{100.0} & \textbf{-0.25} \\
& \texttt{World} & 4 & 0.0 & -1.10 \\
& \textcolor{blue}{\texttt{Sports}} & 5 & \textbf{100.0} & \textbf{-0.02} \\
& \textcolor{blue}{\texttt{Sports}} & 6 & \textbf{100.0} & \textbf{-0.07} \\
\midrule
\multirow{6}{*}{\parbox{3.5cm}{\tiny Article: One way or another, Paul Hamm \#39;s gold-medal odyssey is about to end. Whether he gets to keep the medal and the title he won a month ago in the Olympic men \#39;s gymnastics all-around will be up to the sporting world \#39;s highest authority.}} 
& \textcolor{blue}{\texttt{Sports}} & 1 & \textbf{100.0} & \textbf{-0.05} \\
& \textcolor{blue}{\texttt{Sports}} & 2 & \textbf{100.0} & \textbf{-0.12} \\
& \textcolor{blue}{\texttt{Sports}} & 3 & \textbf{100.0} & \textbf{-0.16} \\
& \texttt{World} & 4 & 0.0 & -1.29 \\
& \textcolor{blue}{\texttt{Sports}} & 5 & \textbf{100.0} & \textbf{-0.03} \\
& \textcolor{blue}{\texttt{Sports}} & 6 & \textbf{100.0} & \textbf{-0.04} \\
\midrule
\multirow{6}{*}{\parbox{3.5cm}{\tiny Article: The role of agents in multimillion-pound football transfer deals came under fresh scrutiny yesterday after Manchester United revealed payments of 11m to middle-men for their help in signing players.}} 
& \textcolor{blue}{\texttt{Sports}} & 1 & \textbf{100.0} & \textbf{-0.28} \\
& \texttt{World} & 2 & 0.0 & -1.03 \\
& \texttt{World} & 3 & 0.0 & -1.32 \\
& \textcolor{blue}{\texttt{Sports}} & 4 & \textbf{100.0} & \textbf{-0.12} \\
& \textcolor{blue}{\texttt{Sports}} & 5 & \textbf{100.0} & \textbf{-0.19} \\
& \textcolor{blue}{\texttt{Sports}} & 6 & \textbf{100.0} & \textbf{-0.16} \\
\midrule
\bottomrule
\end{tabular}%

    }
    \caption{Generated Sequences and Metrics for the AG News Dataset with \texttt{meta/llama-3-8b-instruct}}
\end{table*}

%% file: sst5_appendix.tex
\begin{table*}[h]
    \centering
    \footnotesize 
    \begin{tabular}{c}
        \hline 
        \\
        \begin{minipage}{\textwidth} 
            \begin{itemize}
                \item \textbf{Task Description:}
                \begin{itemize}
                    \item "Classify the sentiment of the following sentences as very negative, negative, neutral, positive, or very positive."
                \end{itemize}

                \item \textbf{Examples:}
                \begin{itemize}
                    \item \textbf{Sentence:} "a 93-minute condensation of a 26-episode tv series, with all of the pitfalls of such you'd expect."
                    \item \textbf{Sentiment:} \texttt{negative}

                    \item \textbf{Sentence:} "this is a startling film that gives you a fascinating, albeit depressing view of iranian rural life close to the iraqi border."
                    \item \textbf{Sentiment:} \texttt{positive}

                    \item \textbf{Sentence:} "but you'll definitely want the t-shirt."
                    \item \textbf{Sentiment:} \texttt{neutral}
                    \end{itemize}
                    \item \textbf{Test Utterance:}
                    \begin{itemize}
                    \item \textbf{Sentence:} "he just wants them to be part of the action, the wallpaper of his chosen reality."
                    \item \textbf{Sentiment:} \texttt{}
                \end{itemize}
            \end{itemize}
        \end{minipage} \\
        \\
        \hline 
    \end{tabular}
    \normalsize 
    \caption{SST-5 Prompt Structure}
    \label{tab:sst5_prompt}
\end{table*}

\begin{table*}[ht]
    \centering
    \small
    \resizebox{\textwidth}{!}{%

\begin{tabular}{p{4.0cm} p{3.5cm} c c c }
\toprule
\textbf{Utterance} & \textbf{Generated Sequence} & \textbf{Order} & \textbf{Accuracy} & \textbf{Log Prob \ours} \\
\midrule
\multirow{6}{*}{\parbox{3.5cm}{\tiny Sentence: a sudsy cautionary tale .}} 
& \textcolor{blue}{\texttt{neutral}} & 1 & \textbf{100.0} & \textbf{-5.69} \\
& \textcolor{blue}{\texttt{neutral}} & 2 & \textbf{100.0} & \textbf{-5.69} \\
& \textcolor{blue}{\texttt{neutral}} & 3 & \textbf{100.0} & \textbf{-5.69} \\
& \textcolor{blue}{\texttt{neutral}} & 4 & \textbf{100.0} & \textbf{-5.69} \\
& \textcolor{blue}{\texttt{neutral}} & 5 & \textbf{100.0} & \textbf{-5.69} \\
& \texttt{negative} & 6 & 0.0 & -66.24 \\
\midrule
\multirow{6}{*}{\parbox{3.5cm}{\tiny Sentence: alex nohe 's documentary plays like a travelogue for what mostly resembles a real-life , big-budget nc-17 version of tank girl .}} 
& \texttt{positive} & 1 & 0.0 & -8.92 \\
& \textcolor{blue}{\texttt{neutral}} & 2 & \textbf{100.0} & \textbf{-6.79} \\
& \texttt{negative} & 3 & 0.0 & -173.75 \\
& \texttt{positive} & 4 & 0.0 & -8.92 \\
& \texttt{positive} & 5 & 0.0 & -8.92 \\
& \texttt{negative} & 6 & 0.0 & -173.75 \\
\midrule
\multirow{6}{*}{\parbox{3.5cm}{\tiny Sentence: here , thankfully , they are .}} 
& \texttt{positive} & 1 & 0.0 & -7.76 \\
& \textcolor{blue}{\texttt{neutral}} & 2 & \textbf{100.0} & \textbf{-5.76} \\
& \texttt{positive} & 3 & 0.0 & -7.76 \\
& \texttt{positive} & 4 & 0.0 & -7.76 \\
& \texttt{positive} & 5 & 0.0 & -7.76 \\
& \texttt{positive} & 6 & 0.0 & -7.76 \\
\midrule
\multirow{6}{*}{\parbox{3.5cm}{\tiny Sentence: hip-hop has a history , and it 's a metaphor for this love story .}} 
& \texttt{positive} & 1 & 0.0 & -5.67 \\
& \textcolor{blue}{\texttt{neutral}} & 2 & \textbf{100.0} & \textbf{-5.55} \\
& \textcolor{blue}{\texttt{neutral}} & 3 & \textbf{100.0} & \textbf{-5.55} \\
& \texttt{positive} & 4 & 0.0 & -5.67 \\
& \texttt{positive} & 5 & 0.0 & -5.67 \\
& \texttt{positive} & 6 & 0.0 & -5.67 \\
\midrule
\multirow{6}{*}{\parbox{3.5cm}{\tiny Sentence: lucas , take notes .}} 
& \texttt{very negative} & 1 & 0.0 & -13.19 \\
& \textcolor{blue}{\texttt{neutral}} & 2 & \textbf{100.0} & \textbf{-5.57} \\
& \texttt{very negative} & 3 & 0.0 & -13.19 \\
& \texttt{positive} & 4 & 0.0 & -8.38 \\
& \texttt{very positive} & 5 & 0.0 & -15.00 \\
& \texttt{very negative} & 6 & 0.0 & -13.19 \\
\midrule
\multirow{6}{*}{\parbox{3.5cm}{\tiny Sentence: taken purely as an exercise in style , this oppressively gloomy techno-horror clambake is impossible to ignore .}} 
& \texttt{positive} & 1 & 0.0 & -9.48 \\
& \textcolor{blue}{\texttt{neutral}} & 2 & \textbf{100.0} & \textbf{-8.01} \\
& \textcolor{blue}{\texttt{neutral}} & 3 & \textbf{100.0} & \textbf{-8.01} \\
& \texttt{positive} & 4 & 0.0 & -9.48 \\
& \texttt{positive} & 5 & 0.0 & -9.48 \\
& \texttt{positive} & 6 & 0.0 & -9.48 \\
\midrule
\multirow{6}{*}{\parbox{3.5cm}{\tiny Sentence: the cartoon that is n't really good enough to be on afternoon tv is now a movie that is n't really good enough to be in theaters .}} 
& \textcolor{blue}{\texttt{very negative}} & 1 & \textbf{100.0} & \textbf{-13.25} \\
& \texttt{negative} & 2 & 0.0 & -113.58 \\
& \texttt{negative} & 3 & 0.0 & -113.58 \\
& \texttt{negative} & 4 & 0.0 & -113.58 \\
& \texttt{negative} & 5 & 0.0 & -113.58 \\
& \texttt{negative} & 6 & 0.0 & -113.58 \\
\midrule
\multirow{6}{*}{\parbox{3.5cm}{\tiny Sentence: the movie 's ripe , enrapturing beauty will tempt those willing to probe its inscrutable mysteries .}} 
& \textcolor{blue}{\texttt{positive}} & 1 & \textbf{100.0} & \textbf{-5.41} \\
& \textcolor{blue}{\texttt{positive}} & 2 & \textbf{100.0} & \textbf{-5.41} \\
& \texttt{very positive} & 3 & 0.0 & -12.19 \\
& \textcolor{blue}{\texttt{positive}} & 4 & \textbf{100.0} & \textbf{-5.41} \\
& \textcolor{blue}{\texttt{positive}} & 5 & \textbf{100.0} & \textbf{-5.41} \\
& \textcolor{blue}{\texttt{positive}} & 6 & \textbf{100.0} & \textbf{-5.41} \\
\midrule
\multirow{6}{*}{\parbox{3.5cm}{\tiny Sentence: two hours of melodramatic musical married to two hours of underdog sports intrigue , if the picture also shares the weaknesses of both genres , more 's the pity .}} 
& \textcolor{blue}{\texttt{neutral}} & 1 & \textbf{100.0} & \textbf{-6.23} \\
& \textcolor{blue}{\texttt{neutral}} & 2 & \textbf{100.0} & \textbf{-6.23} \\
& \texttt{negative} & 3 & 0.0 & -181.70 \\
& \textcolor{blue}{\texttt{neutral}} & 4 & \textbf{100.0} & \textbf{-6.23} \\
& \textcolor{blue}{\texttt{neutral}} & 5 & \textbf{100.0} & \textbf{-6.23} \\
& \texttt{negative} & 6 & 0.0 & -181.70 \\
\midrule
\bottomrule
\end{tabular}%

    }
    \caption{Generated Sequences and Metrics for the SST-5 Dataset with \texttt{meta/llama-3-8b-instruct}}
\end{table*}

%% file: trec_appendix.tex
\begin{table*}[h]
    \centering
    \footnotesize 
    \begin{tabular}{c}
        \hline 
        \\
        \begin{minipage}{\textwidth} 
            \begin{itemize}
                \item \textbf{Task Description:}
                \begin{itemize}
                    \item "Classify the type of the following questions into Abbreviation, Entity, Description, Human, Location, or Number."
                \end{itemize}
                
                \item \textbf{Examples:}
                \begin{itemize}
                    \item \textbf{Question:} "How far is it from Denver to Aspen?"
                    \item \textbf{Type:} \texttt{Number}

                    \item \textbf{Question:} "What county is Modesto, California in?"
                    \item \textbf{Type:} \texttt{Location}

                    \item \textbf{Question:} "Who was Galileo?"
                    \item \textbf{Type:} \texttt{Human}
                    \end{itemize}
                    \item \textbf{Text Utterance:}
                    \begin{itemize}
                    \item \textbf{Question:} "What is the capital of Yugoslavia?"
                    \item \textbf{Type:} \texttt{}
                \end{itemize}
            \end{itemize}
        \end{minipage} \\
        \\
        \hline 
    \end{tabular}
    \normalsize 
    \caption{TREC Prompt Structure}
    \label{tab:trec_prompt}
\end{table*}

\begin{table*}[ht]
    \centering
    \small
    \resizebox{\textwidth}{!}{%

    \begin{tabular}{p{4.0cm} p{3.5cm} c c c }
\toprule
\textbf{Utterance} & \textbf{Generated Sequence} & \textbf{Order} & \textbf{Accuracy} & \textbf{Log Prob \ours} \\
\midrule
\multirow{6}{*}{\parbox{3.5cm}{\tiny Question: How far is it from Denver to Aspen ?}} 
& \textcolor{blue}{\texttt{Number}} & 1 & \textbf{100.0} & \textbf{-5.34} \\
& \textcolor{blue}{\texttt{Number}} & 2 & \textbf{100.0} & \textbf{-5.34} \\
& \textcolor{blue}{\texttt{Number}} & 3 & \textbf{100.0} & \textbf{-5.34} \\
& \texttt{Abbreviation} & 4 & 0.0 & -12.68 \\
& \textcolor{blue}{\texttt{Number}} & 5 & \textbf{100.0} & \textbf{-5.34} \\
& \textcolor{blue}{\texttt{Number}} & 6 & \textbf{100.0} & \textbf{-5.34} \\
\midrule
\multirow{6}{*}{\parbox{3.5cm}{\tiny Question: What city had a world fair in 1900 ?}} 
& \texttt{Entity} & 1 & 0.0 & -9.72 \\
& \textcolor{blue}{\texttt{Location}} & 2 & \textbf{100.0} & \textbf{-7.85} \\
& \texttt{Entity} & 3 & 0.0 & -9.72 \\
& \texttt{Entity} & 4 & 0.0 & -9.72 \\
& \texttt{Entity} & 5 & 0.0 & -9.72 \\
& \textcolor{blue}{\texttt{Location}} & 6 & \textbf{100.0} & \textbf{-7.85} \\
\midrule
\multirow{6}{*}{\parbox{3.5cm}{\tiny Question: What hemisphere is the Philippines in ?}} 
& \texttt{Entity} & 1 & 0.0 & -11.31 \\
& \texttt{Entity} & 2 & 0.0 & -11.31 \\
& \texttt{Entity} & 3 & 0.0 & -11.31 \\
& \textcolor{blue}{\texttt{Location}} & 4 & \textbf{100.0} & \textbf{-7.69} \\
& \texttt{Entity} & 5 & 0.0 & -11.31 \\
& \textcolor{blue}{\texttt{Location}} & 6 & \textbf{100.0} & \textbf{-7.69} \\
\midrule

\multirow{6}{*}{\parbox{3.5cm}{\tiny Question: What is the average weight of a Yellow Labrador ?}} 
& \textcolor{blue}{\texttt{Number}} & 1 & \textbf{100.0} & \textbf{-5.77} \\
& \textcolor{blue}{\texttt{Number}} & 2 & \textbf{100.0} & \textbf{-5.77} \\
& \texttt{Entity} & 3 & 0.0 & -9.98 \\
& \textcolor{blue}{\texttt{Number}} & 4 & \textbf{100.0} & \textbf{-5.79} \\
& \textcolor{blue}{\texttt{Number}} & 5 & \textbf{100.0} & \textbf{-5.77} \\
& \textcolor{blue}{\texttt{Number}} & 6 & \textbf{100.0} & \textbf{-5.77} \\
\midrule
\multirow{6}{*}{\parbox{3.5cm}{\tiny Question: What is the temperature at the center of the earth ?}} 
& \texttt{Description} & 1 & 0.0 & -9.14 \\
& \textcolor{blue}{\texttt{Number}} & 2 & \textbf{100.0} & \textbf{-7.79} \\
& \texttt{Description} & 3 & 0.0 & -9.14 \\
& \texttt{Description} & 4 & 0.0 & -9.14 \\
& \texttt{Description} & 5 & 0.0 & -9.14 \\
& \texttt{Description} & 6 & 0.0 & -9.14 \\
\midrule
\multirow{6}{*}{\parbox{3.5cm}{\tiny Question: What person 's head is on a dime ?}} 
& \textcolor{blue}{\texttt{Human}} & 1 & \textbf{100.0} & \textbf{-7.28} \\
& \texttt{Entity} & 2 & 0.0 & -12.39 \\
& \textcolor{blue}{\texttt{Human}} & 3 & \textbf{100.0} & \textbf{-7.28} \\
& \textcolor{blue}{\texttt{Human}} & 4 & \textbf{100.0} & \textbf{-7.28} \\
& \textcolor{blue}{\texttt{Human}} & 5 & \textbf{100.0} & \textbf{-7.28} \\
& \textcolor{blue}{\texttt{Human}} & 6 & \textbf{100.0} & \textbf{-7.28} \\
\midrule
\multirow{6}{*}{\parbox{3.5cm}{\tiny Question: When did Hawaii become a state ?}} 
& \texttt{Human} & 1 & 0.0 & -9.52 \\
& \textcolor{blue}{\texttt{Number}} & 2 & \textbf{100.0} & \textbf{-7.74} \\
& \texttt{Description} & 3 & 0.0 & -9.07 \\
& \textcolor{blue}{\texttt{Number}} & 4 & \textbf{100.0} & \textbf{-7.74} \\
& \textcolor{blue}{\texttt{Number}} & 5 & \textbf{100.0} & \textbf{-7.74} \\
& \textcolor{blue}{\texttt{Number}} & 6 & \textbf{100.0} & \textbf{-7.74} \\
\midrule
\multirow{6}{*}{\parbox{3.5cm}{\tiny Question: Who developed the vaccination against polio ?}} 
& \textcolor{blue}{\texttt{Human}} & 1 & \textbf{100.0} & \textbf{-10.27} \\
& \textcolor{blue}{\texttt{Human}} & 2 & \textbf{100.0} & \textbf{-10.27} \\
& \texttt{Entity} & 3 & 0.0 & -56.30 \\
& \textcolor{blue}{\texttt{Human}} & 4 & \textbf{100.0} & \textbf{-10.27} \\
& \textcolor{blue}{\texttt{Human}} & 5 & \textbf{100.0} & \textbf{-10.27} \\
& \textcolor{blue}{\texttt{Human}} & 6 & \textbf{100.0} & \textbf{-10.27} \\
\midrule
\multirow{6}{*}{\parbox{3.5cm}{\tiny Question: Who was Galileo ?}} 
& \textcolor{blue}{\texttt{Human}} & 1 & \textbf{100.0} & \textbf{-6.82} \\
& \textcolor{blue}{\texttt{Human}} & 2 & \textbf{100.0} & \textbf{-6.82} \\
& \textcolor{blue}{\texttt{Human}} & 3 & \textbf{100.0} & \textbf{-6.82} \\
& \texttt{Entity} & 4 & 0.0 & -7.44 \\
& \texttt{Entity} & 5 & 0.0 & -7.44 \\
& \textcolor{blue}{\texttt{Human}} & 6 & \textbf{100.0} & \textbf{-6.82} \\
\midrule
\bottomrule
\end{tabular}%

    }
    \caption{Generated Sequences and Metrics for the TREC Dataset with \texttt{meta/llama-3-8b-instruct}}
\end{table*}

%% file: acl_latex.bbl
\begin{thebibliography}{36}
\providecommand{\natexlab}[1]{#1}

\bibitem[{Agarwal et~al.(2024)Agarwal, Singh, Zhang, Bohnet, Rosias, Chan, Zhang, Anand, Abbas, Nova et~al.}]{agarwal2024many}
Rishabh Agarwal, Avi Singh, Lei~M Zhang, Bernd Bohnet, Luis Rosias, Stephanie Chan, Biao Zhang, Ankesh Anand, Zaheer Abbas, Azade Nova, et~al. 2024.
\newblock Many-shot in-context learning.
\newblock \emph{arXiv preprint arXiv:2404.11018}.

\bibitem[{Brown et~al.(2020)Brown, Mann, Ryder, Subbiah, Kaplan, Dhariwal, Neelakantan, Shyam, Sastry, Askell et~al.}]{brown2020language}
Tom Brown, Benjamin Mann, Nick Ryder, Melanie Subbiah, Jared~D Kaplan, Prafulla Dhariwal, Arvind Neelakantan, Pranav Shyam, Girish Sastry, Amanda Askell, et~al. 2020.
\newblock Language models are few-shot learners.
\newblock \emph{Advances in neural information processing systems}, 33:1877--1901.

\bibitem[{Chandra et~al.(2024)Chandra, Ganguly, and Ounis}]{chandra2024predicting}
Manish Chandra, Debasis Ganguly, and Iadh Ounis. 2024.
\newblock One size doesn’t fit all: Predicting the number of examples for in-context learning.

\bibitem[{Chen et~al.(2021)Chen, Zhong, Zha, Karypis, and He}]{chen2022meta}
Yanda Chen, Ruiqi Zhong, Sheng Zha, George Karypis, and He~He. 2021.
\newblock Meta-learning via language model in-context tuning.
\newblock \emph{arXiv preprint arXiv:2110.07814}.

\bibitem[{Guo et~al.(2024)Guo, Wang, Wang, Ye, and Zhang}]{guo2024makes}
Qi~Guo, Leiyu Wang, Yidong Wang, Wei Ye, and Shikun Zhang. 2024.
\newblock What makes a good order of examples in in-context learning.
\newblock In \emph{Findings of the Association for Computational Linguistics: ACL 2024}, pages 14892--14904.

\bibitem[{Gupta et~al.(2023)Gupta, Gardner, and Singh}]{gupta2023coverage}
Shivanshu Gupta, Matt Gardner, and Sameer Singh. 2023.
\newblock Coverage-based example selection for in-context learning.
\newblock \emph{arXiv preprint arXiv:2305.14907}.

\bibitem[{Hovy et~al.(2000)Hovy, Gerber, Hermjakob, Junk, and Lin}]{hovy2001question}
Eduard~H Hovy, Laurie Gerber, Ulf Hermjakob, Michael Junk, and Chin-Yew Lin. 2000.
\newblock Question answering in webclopedia.
\newblock In \emph{TREC}, volume~52, pages 53--56.

\bibitem[{IBM(2023)}]{ibm2023granite}
IBM. 2023.
\newblock Granite: Scaling language models with ibm's efficient architecture.
\newblock \emph{IBM Research Journal}.
\newblock Available at \url{https://research.ibm.com/granite}.

\bibitem[{Liu et~al.(2024{\natexlab{a}})Liu, Liu, Huang, Zhan, Sun, Deng, Wei, and Zhang}]{liu2024se2}
Haoyu Liu, Jianfeng Liu, Shaohan Huang, Yuefeng Zhan, Hao Sun, Weiwei Deng, Furu Wei, and Qi~Zhang. 2024{\natexlab{a}}.
\newblock se2: Sequential example selection for in-context learning.
\newblock In \emph{Findings of the Association for Computational Linguistics ACL 2024}, pages 5262--5284.

\bibitem[{Liu et~al.(2021)Liu, Shen, Zhang, Dolan, Carin, and Chen}]{liu2021makes}
Jiachang Liu, Dinghan Shen, Yizhe Zhang, Bill Dolan, Lawrence Carin, and Weizhu Chen. 2021.
\newblock What makes good in-context examples for gpt-$3 $?
\newblock \emph{arXiv preprint arXiv:2101.06804}.

\bibitem[{Liu et~al.(2024{\natexlab{b}})Liu, Lin, Hewitt, Paranjape, Bevilacqua, Petroni, and Liang}]{liu2024lost}
Nelson~F Liu, Kevin Lin, John Hewitt, Ashwin Paranjape, Michele Bevilacqua, Fabio Petroni, and Percy Liang. 2024{\natexlab{b}}.
\newblock Lost in the middle: How language models use long contexts.
\newblock \emph{Transactions of the Association for Computational Linguistics}, 12:157--173.

\bibitem[{Liu et~al.(2023)Liu, Yuan, Fu, Jiang, Hayashi, and Neubig}]{liu2023pre}
Pengfei Liu, Weizhe Yuan, Jinlan Fu, Zhengbao Jiang, Hiroaki Hayashi, and Graham Neubig. 2023.
\newblock Pre-train, prompt, and predict: A systematic survey of prompting methods in natural language processing.
\newblock \emph{ACM Computing Surveys}, 55(9):1--35.

\bibitem[{Liu et~al.(2024{\natexlab{c}})Liu, Liu, Shi, Cheng, Huang, and Lu}]{liu2024curriculum}
Yinpeng Liu, Jiawei Liu, Xiang Shi, Qikai Cheng, Yong Huang, and Wei Lu. 2024{\natexlab{c}}.
\newblock Let's learn step by step: Enhancing in-context learning ability with curriculum learning.
\newblock \emph{arXiv preprint arXiv:2402.10738}.

\bibitem[{Long et~al.(2024)Long, Zhao, Brown, Xie, Zhao, Chen, Kawaguchi, Shieh, and He}]{do2023adversarial_icl}
Do~Long, Yiran Zhao, Hannah Brown, Yuxi Xie, James Zhao, Nancy Chen, Kenji Kawaguchi, Michael Shieh, and Junxian He. 2024.
\newblock Prompt optimization via adversarial in-context learning.
\newblock In \emph{Proceedings of the 62nd Annual Meeting of the Association for Computational Linguistics (Volume 1: Long Papers)}, pages 7308--7327.

\bibitem[{Lu et~al.(2021)Lu, Bartolo, Moore, Riedel, and Stenetorp}]{lu2021fantastically}
Yao Lu, Max Bartolo, Alastair Moore, Sebastian Riedel, and Pontus Stenetorp. 2021.
\newblock Fantastically ordered prompts and where to find them: Overcoming few-shot prompt order sensitivity.
\newblock \emph{arXiv preprint arXiv:2104.08786}.

\bibitem[{MistralAI(2023)}]{mistral2023mixtral}
MistralAI. 2023.
\newblock Mixtral: A diverse and scalable instruction-tuned language model.
\newblock \emph{Mistral AI Technical Report}.
\newblock Available at \url{https://mistral.ai/mixtral}.

\bibitem[{Perez et~al.(2021)Perez, Kiela, and Cho}]{perez2021truefewshotlearninglanguag}
Ethan Perez, Douwe Kiela, and Kyunghyun Cho. 2021.
\newblock True few-shot learning with language models.
\newblock \emph{Advances in neural information processing systems}, 34:11054--11070.

\bibitem[{Purohit et~al.(2024)Purohit, Devalla, Yerragorla, Bhattacharya, Anand et~al.}]{purohit2024explora}
Kiran Purohit, Raghuram Devalla, Krishna~Mohan Yerragorla, Sourangshu Bhattacharya, Avishek Anand, et~al. 2024.
\newblock Explora: Efficient exemplar subset selection for complex reasoning.
\newblock \emph{arXiv preprint arXiv:2411.03877}.

\bibitem[{Qin et~al.(2023)Qin, Liang, Ye, Zhu, Yan, Lu, Lin, Cong, Tang, Qian et~al.}]{lu2022toolbench}
Yujia Qin, Shihao Liang, Yining Ye, Kunlun Zhu, Lan Yan, Yaxi Lu, Yankai Lin, Xin Cong, Xiangru Tang, Bill Qian, et~al. 2023.
\newblock Toolllm: Facilitating large language models to master 16000+ real-world apis.
\newblock \emph{arXiv preprint arXiv:2307.16789}.

\bibitem[{Radford et~al.(2019)Radford, Wu, Child, Luan, Amodei, Sutskever et~al.}]{radford2019language}
Alec Radford, Jeffrey Wu, Rewon Child, David Luan, Dario Amodei, Ilya Sutskever, et~al. 2019.
\newblock Language models are unsupervised multitask learners.
\newblock \emph{OpenAI blog}, 1(8):9.

\bibitem[{Raffel et~al.(2020)Raffel, Shazeer, Roberts, Lee, Narang, Matena, Zhou, Li, and Liu}]{raffel2020exploring}
Colin Raffel, Noam Shazeer, Adam Roberts, Katherine Lee, Sharan Narang, Michael Matena, Yanqi Zhou, Wei Li, and Peter~J Liu. 2020.
\newblock Exploring the limits of transfer learning with a unified text-to-text transformer.
\newblock \emph{Journal of machine learning research}, 21(140):1--67.

\bibitem[{Reimers(2019)}]{reimers2019sentence}
N~Reimers. 2019.
\newblock Sentence-bert: Sentence embeddings using siamese bert-networks.
\newblock \emph{arXiv preprint arXiv:1908.10084}.

\bibitem[{Shin et~al.(2020)Shin, Razeghi, Logan~IV, Wallace, and Singh}]{shin2020autoprompt}
Taylor Shin, Yasaman Razeghi, Robert~L Logan~IV, Eric Wallace, and Sameer Singh. 2020.
\newblock Autoprompt: Eliciting knowledge from language models with automatically generated prompts.
\newblock \emph{arXiv preprint arXiv:2010.15980}.

\bibitem[{Socher et~al.(2013)Socher, Perelygin, Wu, Chuang, Manning, Ng, and Potts}]{socher2013recursive}
Richard Socher, Alex Perelygin, Jean Wu, Jason Chuang, Christopher~D Manning, Andrew~Y Ng, and Christopher Potts. 2013.
\newblock Recursive deep models for semantic compositionality over a sentiment treebank.
\newblock In \emph{Proceedings of the 2013 conference on empirical methods in natural language processing}, pages 1631--1642.

\bibitem[{Song et~al.(2023)Song, Xiong, Zhu, Wu, Qian, Song, Huang, Li, Wang, Yao et~al.}]{wu2023restgpt}
Yifan Song, Weimin Xiong, Dawei Zhu, Wenhao Wu, Han Qian, Mingbo Song, Hailiang Huang, Cheng Li, Ke~Wang, Rong Yao, et~al. 2023.
\newblock Restgpt: Connecting large language models with real-world restful apis.
\newblock \emph{arXiv preprint arXiv:2306.06624}.

\bibitem[{Sorensen et~al.(2022)Sorensen, Robinson, Rytting, Shaw, Rogers, Delorey, Khalil, Fulda, and Wingate}]{sorensen2022information}
Taylor Sorensen, Joshua Robinson, Christopher~Michael Rytting, Alexander~Glenn Shaw, Kyle~Jeffrey Rogers, Alexia~Pauline Delorey, Mahmoud Khalil, Nancy Fulda, and David Wingate. 2022.
\newblock An information-theoretic approach to prompt engineering without ground truth labels.
\newblock \emph{arXiv preprint arXiv:2203.11364}.

\bibitem[{Touvron et~al.(2023)Touvron, Lavril, Izacard, Martinet, Lachaux, Lacroix, Rozi{\`e}re, Goyal, Hambro, Azhar et~al.}]{touvron2023llama3}
Hugo Touvron, Thibaut Lavril, Gautier Izacard, Xavier Martinet, Marie-Anne Lachaux, Timoth{\'e}e Lacroix, Baptiste Rozi{\`e}re, Naman Goyal, Eric Hambro, Faisal Azhar, et~al. 2023.
\newblock Llama 3: Open and efficient foundation language models.
\newblock \emph{arXiv preprint arXiv:2307.09288}.

\bibitem[{Wu et~al.(2022)Wu, Wang, Ye, and Kong}]{wu2022self_adaptive}
Zhiyong Wu, Yaoxiang Wang, Jiacheng Ye, and Lingpeng Kong. 2022.
\newblock Self-adaptive in-context learning: An information compression perspective for in-context example selection and ordering.
\newblock \emph{arXiv preprint arXiv:2212.10375}.

\bibitem[{Xu et~al.(2024)Xu, Cohen, Wang, and Srikumar}]{xu2024label_distributions}
Zhichao Xu, Daniel Cohen, Bei Wang, and Vivek Srikumar. 2024.
\newblock \href {https://arxiv.org/abs/2402.11447} {In-context example ordering guided by label distributions}.
\newblock \emph{Preprint}, arXiv:2402.11447.

\bibitem[{Yang et~al.(2023)Yang, Zhang, Sui, Liu, Zhao, and Liu}]{yang-etal-2023-representative}
Zhao Yang, Yuanzhe Zhang, Dianbo Sui, Cao Liu, Jun Zhao, and Kang Liu. 2023.
\newblock Representative demonstration selection for in-context learning with two-stage determinantal point process.
\newblock In \emph{Proceedings of the 2023 Conference on Empirical Methods in Natural Language Processing}, pages 5443--5456.

\bibitem[{Ye et~al.(2023)Ye, Wu, Feng, Yu, and Kong}]{ye2023compositional}
Jiacheng Ye, Zhiyong Wu, Jiangtao Feng, Tao Yu, and Lingpeng Kong. 2023.
\newblock Compositional exemplars for in-context learning.
\newblock In \emph{International Conference on Machine Learning}, pages 39818--39833. PMLR.

\bibitem[{Zhang et~al.(2024)Zhang, Lv, Chen, Ha, Xu, and Yan}]{zhang2024batch_icl}
Kaiyi Zhang, Ang Lv, Yuhan Chen, Hansen Ha, Tao Xu, and Rui Yan. 2024.
\newblock Batch-icl: Effective, efficient, and order-agnostic in-context learning.
\newblock \emph{arXiv preprint arXiv:2401.06469}.

\bibitem[{Zhang et~al.(2015)Zhang, Zhao, and LeCun}]{zhang2015character}
Xiang Zhang, Junbo Zhao, and Yann LeCun. 2015.
\newblock Character-level convolutional networks for text classification.
\newblock \emph{Advances in neural information processing systems}, 28.

\bibitem[{Zhang et~al.(2023)Zhang, Zhou, and Liu}]{zhang2023makes}
Yuanhan Zhang, Kaiyang Zhou, and Ziwei Liu. 2023.
\newblock What makes good examples for visual in-context learning?
\newblock \emph{Advances in Neural Information Processing Systems}, 36:17773--17794.

\bibitem[{Zhao et~al.(2024)Zhao, Andriushchenko, Croce, and Flammarion}]{zhao2024instruction_following}
Hao Zhao, Maksym Andriushchenko, Francesco Croce, and Nicolas Flammarion. 2024.
\newblock Is in-context learning sufficient for instruction following in llms?
\newblock \emph{arXiv preprint arXiv:2405.19874}.

\bibitem[{Zhao et~al.(2021)Zhao, Wallace, Feng, Klein, and Singh}]{zhao2021calibrate}
Zihao Zhao, Eric Wallace, Shi Feng, Dan Klein, and Sameer Singh. 2021.
\newblock Calibrate before use: Improving few-shot performance of language models.
\newblock In \emph{International conference on machine learning}, pages 12697--12706. PMLR.

\end{thebibliography}
